\documentclass[twoside]{article}
\usepackage[preprint]{aistats2027}
\usepackage[utf8]{inputenc} 
\usepackage[T1]{fontenc}    
\usepackage{hyperref}       
\usepackage{url}
\usepackage{xurl}            
\usepackage{booktabs}       
\usepackage{amsfonts}       
\usepackage{nicefrac}       
\usepackage{microtype}      
\usepackage{xcolor}   
\usepackage[backend=biber]{biblatex}
\usepackage{amsmath,enumitem,amsfonts}
\usepackage{mathtools}
\usepackage{subcaption}
\usepackage{caption}
\usepackage{comment}
\usepackage{float}
\usepackage{graphicx}
\usepackage{wrapfig}
\usepackage{comment}

\usepackage{amsthm}
\usepackage{bbm}
\usepackage{tikz}
\usetikzlibrary{arrows.meta}

\newcommand{\RR}{\mathbb{R}}

\makeatletter
\newtheorem*{rep@theorem}{\rep@title}
\newcommand{\newreptheorem}[2]{%
\newenvironment{rep#1}[1]{%
 \def\rep@title{#2 \ref{##1}}%
 \begin{rep@theorem}}%
 {\end{rep@theorem}}}
\makeatother

\newtheorem{theorem}{Theorem}[section]
\newtheorem{lemma}{Lemma}[section]

\newtheorem{proposition}{Proposition}[section]
\newtheorem{definition}{Definition}[section]
\newreptheorem{theorem}{Theorem}
\newtheorem{assumption}{Assumption}[section]
\theoremstyle{remark}
\newtheorem{remark}{Remark}[section]

\begin{document}
\twocolumn[
\aistatstitle{The Clustering Occupation Kernel (CLOCK) Method for Time Series Arising from Dynamical Systems}
\aistatsauthor{Michael L. Wells$^*$ 
    \AND
     Kamel Lahouel$^\dagger$ 
     \And
      Victor Rielly$^\dagger$ 
     \AND Bruno Jedynak$^*$ }
\aistatsaddress{
   $^*$Fariborz Maseeh Department of Mathematics and Statistics\\    Portland State University\\
    Portland, OR 97201 \\
    \\
   $^\dagger$Translational Genomics Research Institute \\
    Phoenix, AZ 85004 \\
    \\
   \texttt{\{mlwells, bjedyna2\}@pdx.edu}, \texttt{\{klahouel, vrielly\}@tgen.org}
    }
\runningauthor{Wells, Lahouel, Rielly, Jedynak}
\runningtitle{The Clustering Occupation Kernel (CLOCK) Method}

]


\begin{abstract}
We introduce the Clustering Occupation Kernel (CLOCK) method, a novel representation-based framework for clustering time series arising from dynamical systems. CLOCK embeds observed sequences into a reproducing kernel Hilbert space (RKHS) via a time-averaged occupation kernel, enabling trajectory comparison through their RKHS norm. For systems governed by differential equations, this formulation rigorously approximates the kernel mean embedding of the invariant Sinai-Ruelle-Bowen (SRB) measure, meaning our dissimilarity metric approximates the maximum mean discrepancy (MMD) between attractors. To accommodate general time series, we introduce time lags via Takens’ Embedding Theorem to reconstruct the phase space. Crucially, CLOCK natively handles trajectories of unequal lengths and irregularly spaced observations. We validate the proposed metric using spectral clustering, demonstrating superior performance against benchmark methods on both synthetic datasets and a real-world intracranial large-scale electroencephalogram (EEG) dataset. Finally, ablation studies on the Lorenz-96 system are consistent with the theoretical convergence rates of the occupation kernels and highlight the method's strong robustness to zero-mean Gaussian noise.
\end{abstract}
\section{Introduction}
Time series, sequences of observations with accompanying timestamps \cite{nielsen2019practical}, are ubiquitous objects in data science.  The immense volume of available data makes annotating each time series infeasible, so unsupervised methods are needed to cluster data by shared characteristics.  Time series clustering methods fall roughly into three categories: distance-based, model-based, and representation-based.  Distance-based methods use a custom distance function defined on time series, so standard methods like K-means \cite{mcqueen1967some} can be adapted.  The DTW barycenter-averaging method (DBA) \cite{petitjean2011global} using dynamic time warping \cite{1163055}\cite{berndt1994using} is an example.  
Model-based methods cluster trajectories based on their similarity in the parameters of a model assumed to govern the data.  For example, time-series-based Gaussian mixture models 
fall into this category \cite{UMATANI2023109375}.  Representation-based methods map the time series to latent feature vectors that capture the essential elements of each time series.  Typical clustering methods in Euclidean space may be applied to these representations.  Autoencoders \cite{srivastava2015unsupervised}\cite{malhotra2017timenet} and contrastive-learning-based methods \cite{yue2022ts2vec} are examples.

In this work, we adopt a representation-based perspective that views time series, after preprocessing, as solutions to ordinary differential equations (ODEs).
Specifically, we embed each observed trajectory $y : [0,T] \to M$ into a reproducing kernel Hilbert space (RKHS) by considering its time-averaged occupation embedding \cite{Rosenfeld_2019,rielly2025mock}:
\begin{equation}
\label{eq:Embedding}
    y \;\mapsto\; \frac{1}{T} \int_0^T K(\cdot, y(t))\, dt,
\end{equation}
where $K$ is a positive-definite kernel defined on the state space $M$.

This representation captures the time-averaged occupation measure of the trajectory in state space and provides a natural way to compare time series through Hilbert space distances. Its relevance becomes particularly clear when the observed trajectories are generated by dynamical systems. According to the Palis conjecture \cite{palis2000global}, a generic dynamical system defined as the flow of a $C^r$ ($r \geq 1$) map on a compact manifold admits (i) finitely many attractors, (ii) each of which is supported by a Sinai--Ruelle--Bowen (SRB) measure, and (iii) the union of the basins of these attractors has full Lebesgue measure. 
These SRB measures characterize the long-term statistical behavior of trajectories. 

The main contribution of this work is to (1) show that the embedding defined by \eqref{eq:Embedding} approximates the kernel mean embedding (KME)\cite{sriperumbudur2011universality} of these SRB measures, (2) leverage this connection to propose the Clustering Occupation Kernel method (CLOCK): a principled clustering method that groups trajectories according to their associated SRB measures, and (3) evaluate CLOCK against the state of the art. 

We compute dissimilarities between time series by approximating the maximum mean discrepancy (MMD) between their estimated SRB measures. The authors in \cite{das2021kernel} took a related approach, embedding time series in high-dimensional feature space using sliding windows. Their goal was to provide a kernel-based distance measure that estimates distances by matching feature distributions that only partially overlap in their support. However, for time series generated by a dynamical system, their method clusters trajectories based on transient dynamics governed by the initial distribution. By contrast, our proposed method clusters trajectories according to their long-term, asymptotic statistics, making the resulting distances invariant to initial conditions as long as the trajectories are long enough. Ultimately, the choice between these methods depends on whether the application demands sensitivity to short-term transients or long-term systemic invariants.

Whether the time series is generated by an unknown discrete-time system or a continuous higher-order ODE, we can reconstruct its multi-dimensional phase space by preprocessing the observations using time-delay embedding before using CLOCK. Provided the underlying states evolve on a smooth, compact manifold, Takens' theorem \cite{takens2006detecting} guarantees that these delay-coordinate trajectories provide a diffeomorphism to the true orbits of the underlying system.

We demonstrate the proposed metric's effectiveness through a series of clustering tasks. Specifically, we apply spectral clustering to similarity matrices derived from the MMD distances between trajectory embeddings. We present numerical experiments in which the time series are the solutions of ODEs corrupted by noise. Different clusters correspond to different vector-field parameters. In this context, we compare CLOCK with standard benchmark approaches. We also evaluate CLOCK on a large medical dataset of electroencephalogram measurements with a gold standard. 
\section{Method}
\subsection{Problem setup}
We are given a collection of time series $\{y_i\}_{i=1}^n$, each composed of a finite collection of points $\{y_{ij}\}_{j=0}^{m_i}$ and a corresponding collection of timestamps $\{t_{ij}\}_{j=0}^{m_i}$.  In order to motivate the method, let the observations be noisy snapshots from the solutions to an initial value problem: 
\begin{equation}
\label{eq:ODE_eq}
    y_{ij} = x_i(t_{ij}) + \epsilon_{ij}, \frac{dx_i}{dt} = f_c(x_i), \;\; x_i(t_{i0}) = x_{i0},
\end{equation}
where $c \in \{1,\ldots,C\}, \epsilon_{ij} \sim N(0, \Sigma)$ are i.i.d. and
where $\{f_c\}_{c \in \{1,\ldots, C\}}$ is a finite collection of vector fields, typically within the same family but with different parameters.  The goal is to find a cluster assignment $\{1, \ldots, n\} \to \{1, \ldots, C\}$ that maps each time series 
in the collection $\{(t_{ij}, y_{ij})\}, 1\leq i \leq n, 0 \leq j \leq m_i$ to one of $C$ clusters
without knowledge of the vector fields or the initial conditions. In this context, one could consider a model-based approach jointly estimating the cluster assignment, the vector fields, and the initial conditions using an Expectation-Maximization algorithm \cite{dempster1977maximum}. We present below a more direct approach: computing a similarity between trajectories that can be used as input to an existing clustering algorithm. Note that we do not assume that the time series are of equal length or that the observations are regularly spaced. In fact, the algorithm described below applies to any time-series clustering problem. However, it is justified for data generated by a dynamical system or data that can be preprocessed to exhibit the same statistical characteristics as if it were generated by a dynamical system.   

\subsection{Background}
\label{sec:background}

Let $M$ be a compact manifold, and let $f \colon M \to M$ be a $C^r$ map ($r\geq 1$).  A {\em Sinai--Ruelle--Bowen (SRB) measure for $f$} is an $f$-invariant Borel probability measure $\mu$ on $M$ with several important properties.  We describe the property of interest as follows:  \footnote{Another property is the existence of Lebesgue-absolutely-continuous conditional measures on unstable manifolds.  See \cite{alves2020nonuniformly} for details.}   Let $\phi \colon M \to \mathbb{R}$ be a continuous function and let $y \colon [0, \infty) \to M$ be a solution to the differential equation defined by $f$.  There exists a positive-Lebesgue-measure set, denoted $B(\mu)$ and referred to as the {\em basin of attraction for $\mu$}, such that if $y(0) \in B(\mu)$, then the following holds:
\begin{equation}
    \label{eq:SRB_measure}
    \lim_{T \to \infty}\frac{1}{T}\int_0^T \phi(y(t)) dt = \int_M \phi(p) d\mu(p).
\end{equation}
The Palis conjecture states that, for a dense set $\Phi$ in the space of all $C^r$ ($r \geq 1$) maps $M \to M$ with the $C^r$ topology, each $f \in \Phi$ has a finite collection of SRB measures $\mu_1,\ldots, \mu_n$ such that $\cup_i B(\mu_i)$ has total Lebesgue measure.  That is, $\lambda(\cup_i B(\mu_i)) = \lambda(M)$ where $\lambda(\cdot)$ denotes the Lebesgue measure.

Let $H$ be a reproducing kernel Hilbert space (RKHS) of functions defined on a topological space $\mathcal{X}$. Let $\nu$ be a probability measure on $\mathcal{X}$.  Then the KME of $\nu$ is given by the image of the mapping:
\begin{equation}
\label{eq: KME mappring}
    \nu \mapsto \int_M K(\cdot, p)d\nu(p).
\end{equation}
The 
MMD  between two probability measures $\mu_1, \mu_2$ defined on a topological space is the following quantity:
\begin{equation}
    \text{\rm MMD}(\mu_1, \mu_2) = \|\hat{\mu}_1 - \hat{\mu}_2\|_H,
\end{equation}
where $\hat{\mu}_1, \hat{\mu}_2$ are the KME's of $\mu_1, \mu_2$, respectively.  The MMD quantifies the dissimilarity of the measures $\mu_1$ and $\mu_2$. We say that $K$ is {\em characteristic} (see \cite{sriperumbudur2011universality}) when the  mapping \eqref{eq: KME mappring}  is one-to-one.  If $K$ is characteristic, then the measures $\mu_1$ and $\mu_2$ are identical if and only if $\text{\rm MMD}(\mu_1,\mu_2) = 0$.
\subsection{The clustering occupation kernel method (CLOCK)}
\label{sec:CLOCK}

Assume for now that each trajectory $y_i$ is observed continuously over the time interval $[t_{i0},t_{im_i}]$. Let us then embed $y_i$ into an RKHS by computing its time-averaged occupation kernel.  Thus, we select a continuous, positive-definite kernel $K$ with corresponding RKHS $H$ and compute the mapping:  
\begin{equation}
    \label{eq:time_avg_occ_kern}
    y_i \mapsto \frac{1}{T_i}\int_{t_{i0}}^{t_{im_i}}K(\cdot, y_i(t)) dt =: L_i^*,
\end{equation}
where $T_i := t_{im_i} - t_{i0}$.
The motivation for this embedding comes from the following result, proved in Appendix  \ref{sec:theoretical_background}:
\begin{reptheorem}{cor:strong_convergence}
    Let $K$ be a continuous, positive-definite kernel defined on a compact manifold $M$.  Let $y \colon [0, \infty) \to M$ be a solution to a first-order ODE satisfying the Palis conjecture.  Then we have that the following holds almost surely:
    \begin{equation}
        \label{eq:kernel_limit}
        \lim_{T \to \infty} \frac{1}{T}\int_0^T K(\cdot, y(t))dt = \int_M K(\cdot, p)d\mu(p),
    \end{equation}
    where $\mu$ is the SRB measure of the attractor of the underlying dynamical system for which $y(0) \in B(\mu)$. 
\end{reptheorem}

 \textbf{Proof sketch:} We prove this theorem in two steps: first we prove weak convergence, and then we prove convergence of the norm, from which strong convergence follows.  
 The convergence of the norm can be shown using the kernel's Mercer property. 

Now, according to Theorem \ref{cor:strong_convergence}, we can approximate
\begin{equation}
    \label{eq:KME_equation}
    L_i^* \approx \int_M K(\cdot, p) d\mu_i(p) =: \hat{\mu}_i,
\end{equation}
where $\mu_i$ is the SRB measure of the attractor such that $y_i(0) \in B(\mu_i)$.  
Note that $\hat{\mu}_i$ is the KME of $\mu_i$.

We are now ready to propose the dissimilarity $d^2(\cdot, \cdot)$ between time series.  It is  defined by:
\begin{multline}
    \label{eq:distance_metric}
    d^2(y_i, y_j) = \|L_i^* - L_j^*\|^2_H =\\ \frac{1}{T_i^2}\int_{t_{i0}}^{t_{im_i}}\int_{t_{i0}}^{t_{im_i}}K(y_i(t), y_i(s)) ds dt +\\ \frac{1}{T_j^2}\int_{t_{j0}}^{t_{jm_j}}\int_{t_{j0}}^{t_{jm_j}}K(y_j(t), y_j(s)) ds dt -\\ \frac{2}{T_iT_j}\int_{t_{i0}}^{t_{im_i}}\int_{t_{j0}}^{t_{jm_j}}K(y_i(t), y_j(s)) ds dt.
\end{multline}
In practice, the time series are observed at a finite number of time points, thus we estimate the integrals with trapezoid-rule quadratures using the samples $\{y_{ij}\}_{j=0}^{m_i}$ and associated time points $\{t_{ij}\}_{j=0}^{m_i}$, see Appendix \ref{sec:approx_dissimilarity}.  This dissimilarity is an estimate of $\|\hat{\mu}_i - \hat{\mu}_j\|^2_H$, the squared MMD of $\mu_i$ and $\mu_j$.
Next, we convert dissimilarities to similarities in order to perform spectral clustering.  We define the similarity $S(\cdot, \cdot)$ according to:
\begin{equation}
    \label{eq:gaussian_similarity}
    S(y_i, y_j) := \exp\left(-\frac{1}{2\sigma^2}\|L_i^* - L_j^*\|_H^2\right),
\end{equation}
where $\sigma := \mbox{\rm median} \{\|L_i^* - L_j^*\|_H\}_{i \neq j}^n$ \cite{garreau2017large}.  We obtain an $(n, n)$ matrix of similarities which we use as input for a standard spectral clustering algorithm.  Note that other similarities may be derived from the given dissimilarity.  All the experiments in this paper are presented with the similarity in \eqref{eq:gaussian_similarity}. 
\subsection{Rate of convergence}

We prove the following result in Appendix~\ref{sec:theoretical_background}:
\begin{reptheorem}{thm:major_thm_2}
    Let $M \subset \mathbb{R}^d$ be a compact forward-invariant region for the
    flow $\Phi_t$ of a $C^1$ vector field, let $\mu$ be an invariant
    probability measure on $M$, and let $K$ be a continuous positive-definite
    kernel with H\"older feature map. Suppose that time-averaged occupation
    kernels started from $\tilde{X} \sim \mu$ converge to $\hat{\mu}$ at
    mean-square rate $\mathcal{O}(T^{-1})$ (e.g.\ under
    Lemma~\ref{lem:big_lemma}), and that trajectories started from
    $X \sim \nu$ are exponentially asymptotic to trajectories on the
    attractor, with asymptotic phases distributed absolutely continuously
    with respect to $\mu$ (Assumption~\ref{ass:asymptotic_phase}). Then
    \begin{equation}
        \left\|\frac{1}{T}\int_0^T K(\cdot, \Phi_t(X))\, dt
        - \int K(\cdot, p)\, d\mu(p)\right\|_H
        = \mathcal{O}_{\mathbb{P}}(T^{-1/2}).
    \end{equation}
\end{reptheorem}
 
\begin{remark}
\label{rem:applicability_main}
Among the systems of Section~\ref{sec:datasets}, the assumptions of
Theorem~\ref{thm:major_thm_2} hold for the FitzHugh--Nagumo system (hyperbolic
sinks) and the Van der Pol system (hyperbolic limit cycles), with every kernel
used in our experiments; see Appendix~\ref{sec:applicability}. For the
chaotic systems the theorem is not established. The theorem concerns
noise-free trajectories as $T \to \infty$, so it motivates, rather than
certifies, the finite-$T$ experiments.
\end{remark}
We also provide a visualization of this rate in Section~\ref{sec:ablation_studies}.

\subsection{Application to general time series}
\label{sec:Takens_thm}
We use Takens' embedding theorem \cite{takens2006detecting} to extend our setting to more general time series.  We augment the time-series dimension by adding time lags. That is, we use trajectories of the form $(y(t), y(t +\tau), \ldots, y(t + m\tau)) \in \mathbb{R}^{m+1}$ where $m \in \mathbb{N}$ and $\tau > 0$ are fixed.  If $\tau$ and $m$ are chosen appropriately, the time-lagged trajectories will be the images of integral curves under a smooth embedding of a compact manifold into $\mathbb{R}^{m+1}$.  Thus, we may approximate the KMEs of their SRB measures.  

The optimal time-lag $\tau$ may be found by computing the first local minimum of the average mutual information of $(y(t), y(t + \tau))$ \cite{fraser1986independent}, while a large enough embedding dimension $m$ may be selected by minimizing the number of false nearest neighbors of the data points \cite{kennel1992determining}.  
\subsection{Scaling the method to large datasets}
\label{sec:handling_large_datasets}
Like distance-based methods, 
this method requires computing dissimilarities for all possible pairs of time series in the dataset, so, by default,  its complexity scales quadratically with the number of time series $n$. However, by approximating an implicit kernel with an explicit one, e.g., $p$ random Fourier features \cite{rahimi2007random} and using truncated (rank k) singular value decomposition, the time and space complexities are reduced to, resp.,  $O(npk)$ and  $O(k(n + p))$.    
\section{Experiments}
\label{sec:experiments}
 
We compared the CLOCK method to a collection of standard methods on several synthetic datasets as well as a real, medical dataset.\footnote{For an implementation of these experiments, see: \protect\url{https://anonymous.4open.science/r/RKHS-CLOCK-36BD}}.  In addition, we performed ablation studies to determine the convergence rate of the time-averaged occupation kernels and the quality of the clustering as a function of trajectory length.  

All experiments on synthetic data were conducted on a laptop with the Intel Core Ultra 9 Processor 185H, 64 GB RAM, and an NVIDIA GPU with 8GB of memory.  The experiment on real data was conducted on a High-Performance Computing (HPC) cluster using a single NVIDIA A40 GPU with 45 GB RAM.
\subsection{Synthetic datasets}
\label{sec:datasets}

\begin{figure*}
    \centering
    \begin{subfigure}{0.18\linewidth}
          \includegraphics[width=1.0\linewidth]{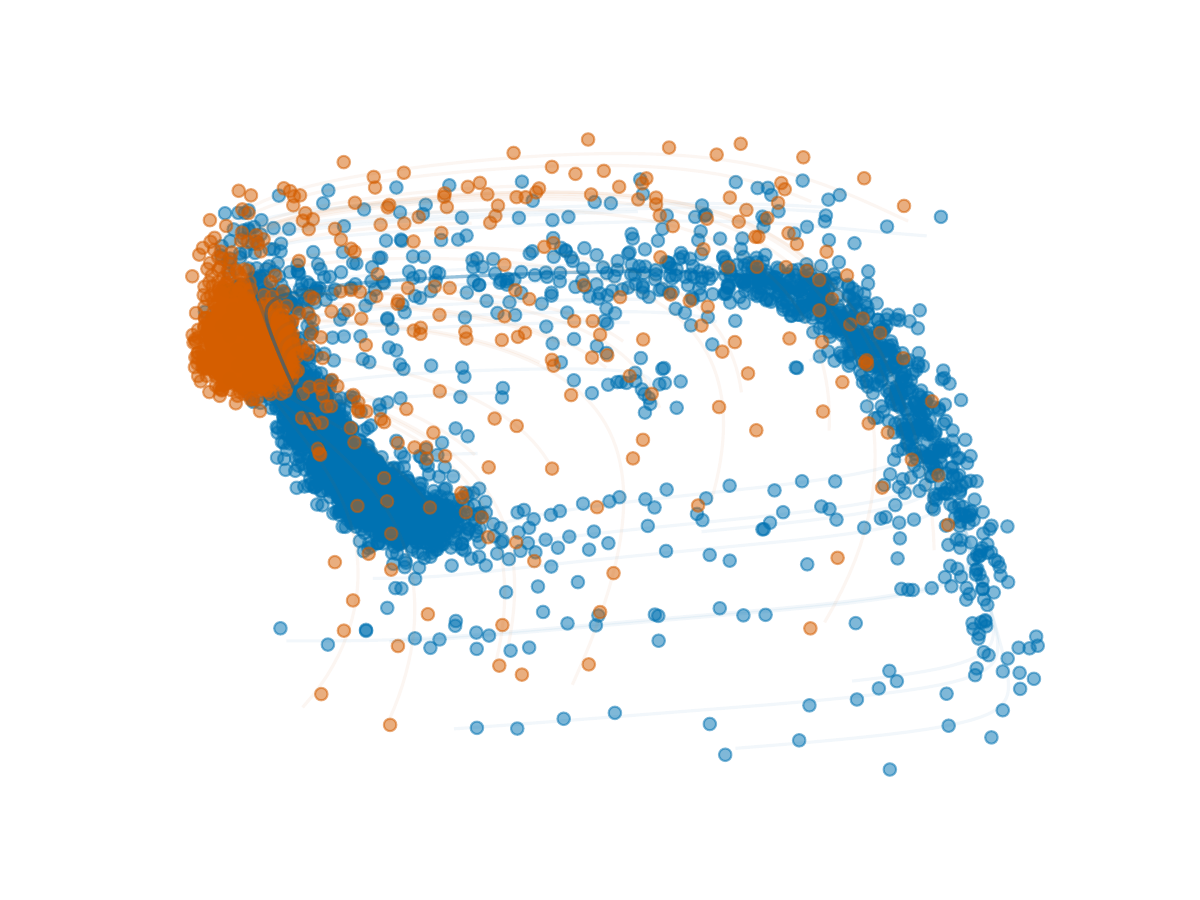}
    \caption{FHN. 
    }
    \label{fig:FHN_dataset}
    \end{subfigure}
\begin{subfigure}{0.18\linewidth}
\includegraphics[width=1.0\linewidth]{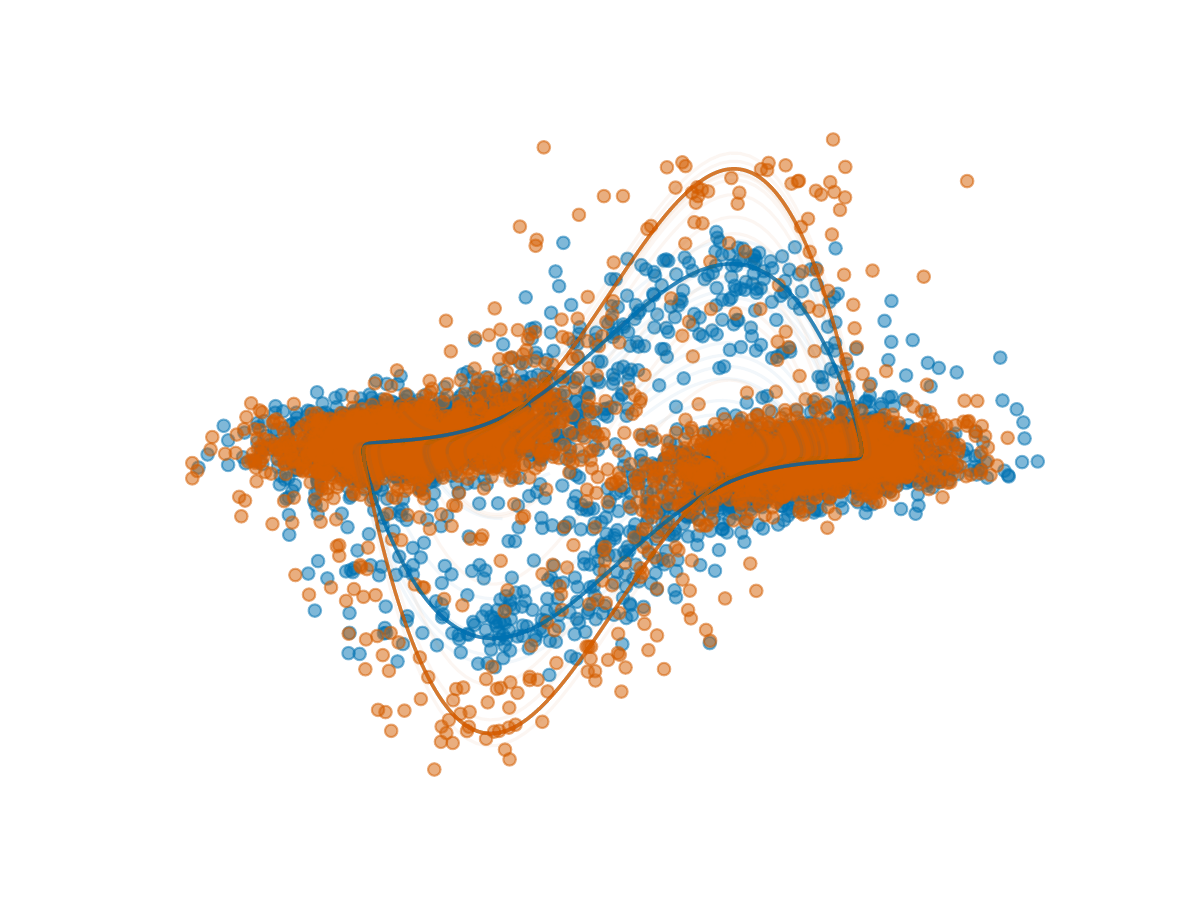}
     \caption{Van der Pol}
      \label{fig:vdp_cluster}
     \end{subfigure}
   \begin{subfigure}{0.18\linewidth}
  
          \includegraphics[width=1.0\linewidth]{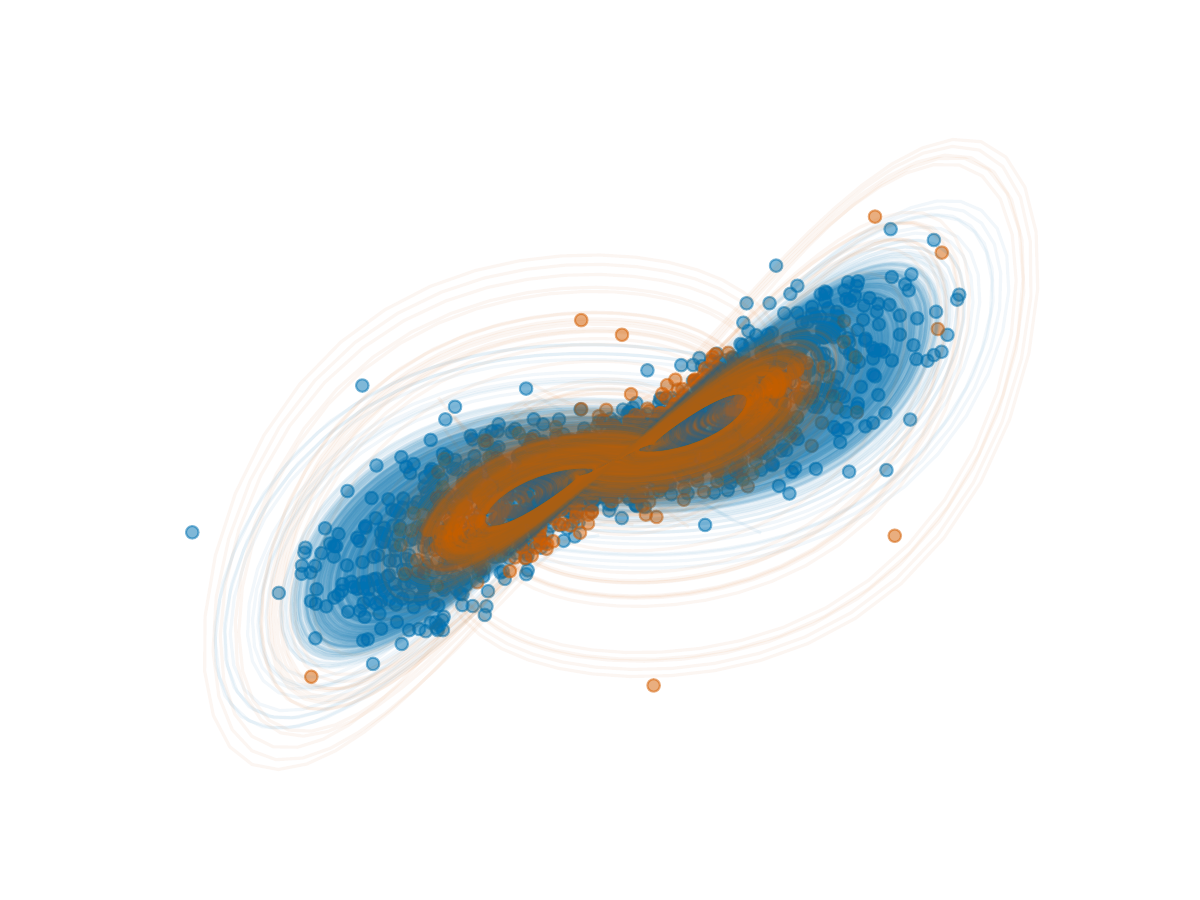}
           \caption{Lorenz63}
    \label{fig:lorenz63}
    \end{subfigure}
\begin{subfigure}{0.18\linewidth}
\includegraphics[width=1.0\linewidth]{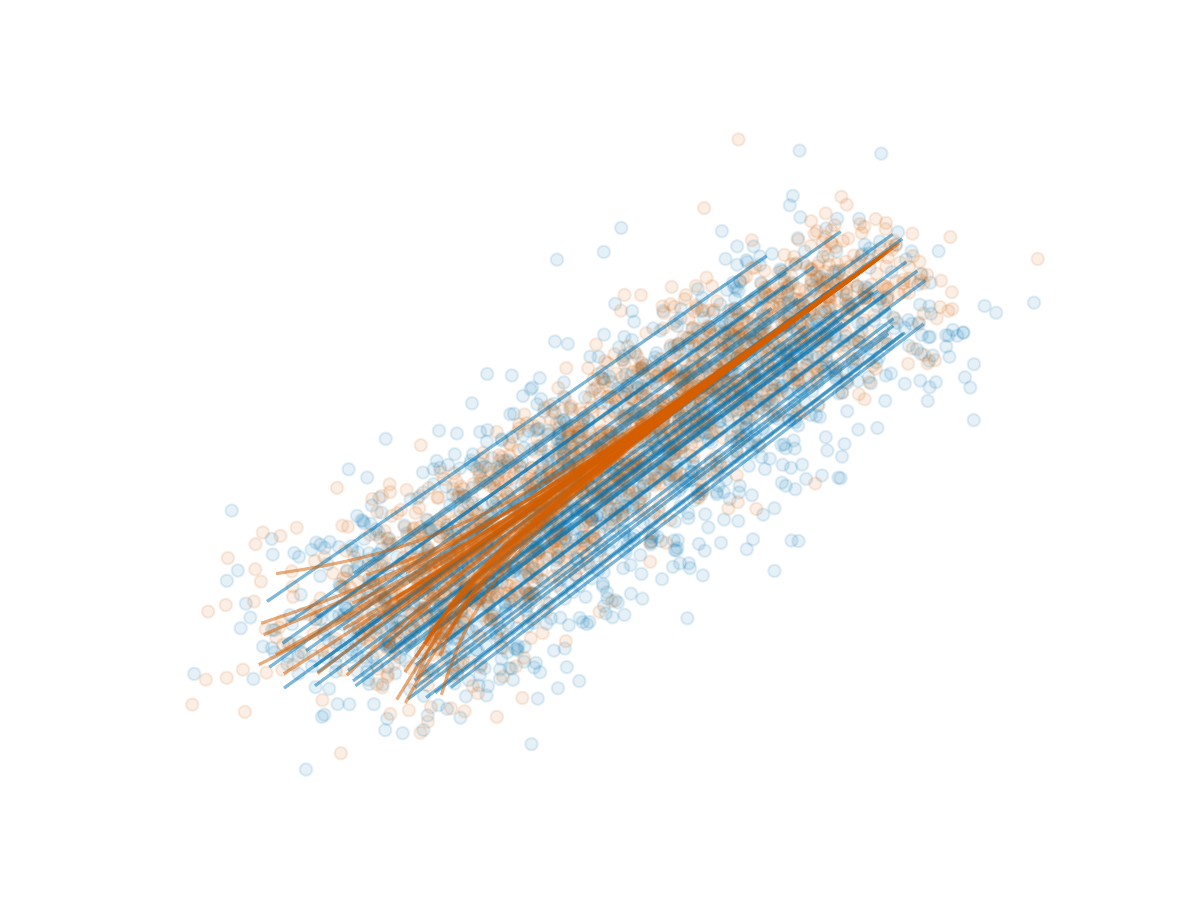}
     \caption{Kuramoto}
      \label{fig:kuramoto}
     \end{subfigure}
     \begin{subfigure}{0.18\linewidth}
\includegraphics[width=1.0\linewidth]{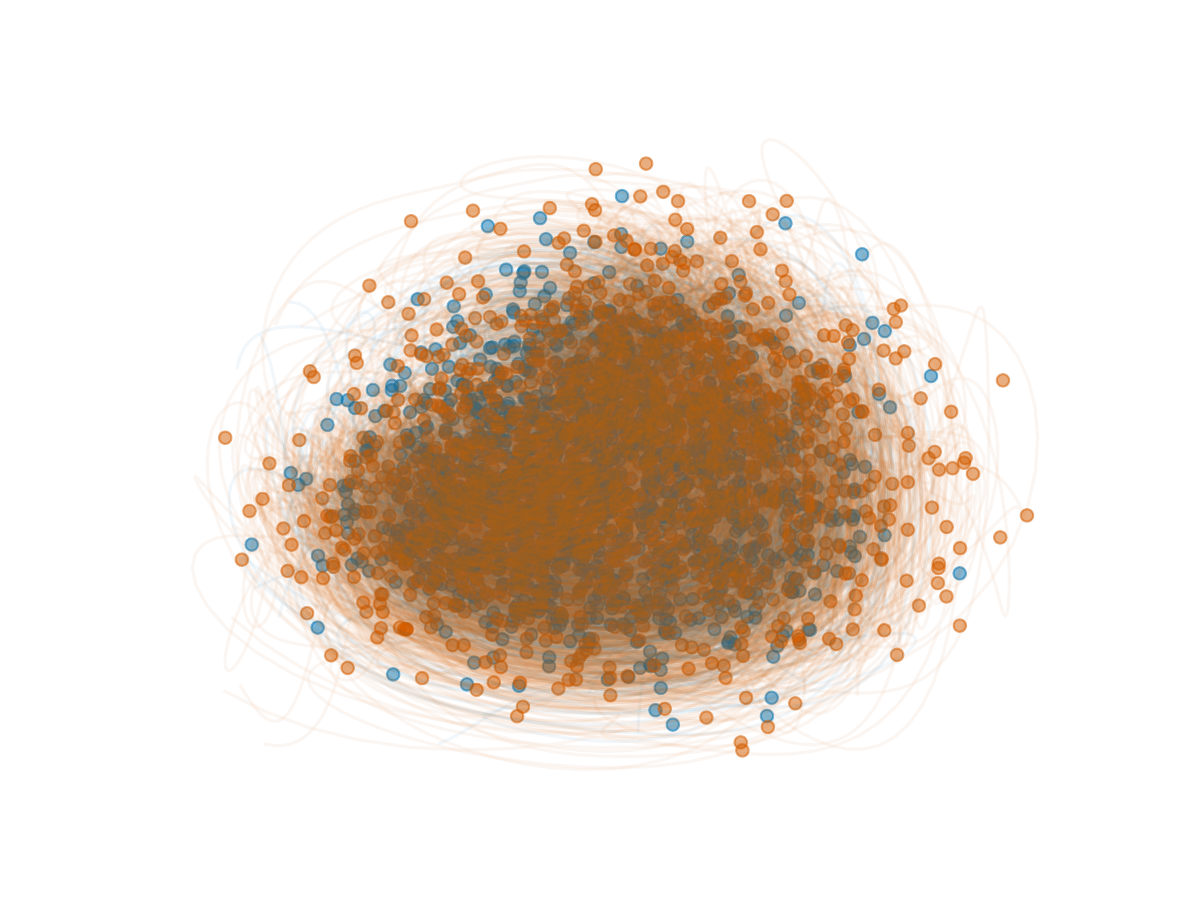}
     \caption{Lorenz96}
      \label{fig:lorenz96}
     \end{subfigure}
    \caption{Shown are the synthetic datasets in the $(x_1, x_2)$-plane with coloring given by the true cluster labels.  We present the true trajectories along with the noisy samples.  For higher-dimensional datapoints, we plot only the first two coordinates.}.
    \label{fig:synthetic_datasets}
\end{figure*}

For the synthetic datasets, we set the number of clusters to be 2 for simplifying the evaluation and generated 30 trajectories per cluster.  Each cluster was generated by the same differential equation but with different parameters.  Trajectories were integrated numerically using a standard fourth-order Runge-Kutta method with a fixed step size. White noise was then added. The datasets are visualized in Figure \ref{fig:synthetic_datasets}.
\begin{enumerate}
    \item {\bf FitzHugh-Nagumo (FHN) oscillator :}\cite{fitzhugh1961impulses}\cite{nagumo2007active}  This is a 2D system given by the equations:
    \begin{multline}
        \frac{dx_1}{dt} = x_1 - \frac{x_1^3}{3} - x_2 + R,\\    \frac{dx_2}{dt} = \frac{1}{\tau}(x_1 + a - bx_2) ,
    \end{multline}
    where $a, b, R, \tau$ are constants.  For cluster 1, we used $a = 0.7, b=0.8, R = 0.2, \tau = 12.5$.  For cluster 2, we used $a=3.2, b=1.7, R=-0.4, \tau = 22.5$.  We sampled initial conditions uniformly from the box $[-2, 2]^2$ and integrated trajectories over the time interval $[0, 20]$.  We selected 101 regularly spaced points on each trajectory.  We added zero-mean Gaussian noise with standard deviation $0.1$.
    \item {\bf Van der Pol oscillator:}\cite{van1926lxxxviii} A 2D system with governing equations:
    \begin{align}
        \frac{dx_1}{dt} &= x_2,\;\;  \frac{dx_2}{dt} = \mu(1-x_1^2)x_2 - x_1,
    \end{align}
    where $\mu$ is a constant.  We selected $\mu= 3$ for cluster 1 and $\mu = 5$ for cluster 2.  Initial conditions were sampled uniformly in the box $[-2, 2]^2$.  We integrated trajectories over the time interval $[0, 80]$.  We then sampled 101 points at irregular intervals on each trajectory.  We added Gaussian noise with standard deviation $0.5$.
    \item {\bf Lorenz63:}\cite{lorenz1963deterministic}  This is a 3D dynamical system.  The governing equations are given by:
    \begin{multline}
        \frac{dx_1}{dt} = \sigma(y-x), \frac{dx_2}{dt} =x(\rho - z) - y,\\ \frac{dx_3}{dt} = xy -\beta z,
    \end{multline}
    where $\sigma, \rho, \beta$ are constants.  For cluster 1, we used $\sigma = 10, \rho = 28, \beta = 8/3$.  For cluster 2, we used $\sigma = 8, \rho = 33, \beta = 2/3$.  Initial conditions were sampled uniformly from the box $[-10, 10]^3$.  We integrated trajectories over the time interval $[0, 60]$ and selected 41 points at irregular intervals on each trajectory.  We then added Gaussian noise with standard deviation $1.5$
    \item {\bf Kuramoto:}\cite{kuramoto2005self}
    We used a 10D Kuramoto model with governing equations:
    \begin{equation}
        \frac{dx_i}{dt} = \omega_i + \frac{\kappa}{10}\sum_{j=1}^N \sin(x_j - x_i)
    \end{equation}
    for $i=1,\ldots,10$ where $\{\omega_i\}_{i=1}^{10}$ and $\kappa$ are constants.  For both clusters, we set $\omega := (1, 1, \ldots, 1, \sqrt{2})$.  For cluster 1, we set $\kappa = 0.05$, and for cluster 2, we set $\kappa = 1$.  Initial conditions were sampled from the box $[-1, 1]^{10}$.  We integrated trajectories over the time interval $[0, 5]$ and selected 51 regularly-spaced points.  We then added Gaussian noise with standard deviation $0.5$.
    \item {\bf Lorenz96:}\cite{lorenz1996predictability}  We used a 40D Lorenz96 system with governing equations:
    \begin{equation}
        \frac{dx_i}{dt} = (x_{i+1} - x_{i-2})x_{i-1} - x_i + F,
    \end{equation}
    where we use the convention $x_{-1} := x_{39}, x_0 := x_{40}$, $x_{41} := x_1$ and $F$ is a constant.  For cluster 1, we used $F = 8$, and for cluster 2, we used $F = 10$.  Initial conditions were sampled uniformly from the box $[-10, 10]^{40}$.  We integrated trajectories over the time interval $[0, 60]$ and selected 41 irregularly-spaced points on each trajectory.  We then added Gaussian noise with standard deviation $1.5$.
\end{enumerate}
\subsection{Real dataset}
\label{sec:real_datasets}
\begin{enumerate}
    \item {\bf Bonn EEG time series dataset:}
    \cite{andrzejak2001indications}
    This dataset consists of 1D intracranial electroencephalogram (EEG) recordings of the human brain used for epileptic seizure detection research.  There were two clusters in the dataset, each with 100 time series.  Cluster 1 consisted of baseline, seizure-free readings in the epileptogenic zone, while Cluster 2 contained measurements of seizure activity taken from all regions of the brain where such activity was found.  Automatically identifying seizure activity in EEG recordings has the potential to provide insight into brain function, and to aid in diagnosis and seizure prevention.  Each time series was 23.6 seconds long with 4097 equally-spaced samples.
    
    For the CLOCK method, we used Takens' embedding theorem to augment the time series (see section \ref{sec:Takens_thm}).
    We set $m=6$ and $\tau= 13$ time-steps for this experiment.  
\end{enumerate}
\subsection{Evaluation metrics}
\label{sec:eval_metrics}
We used the class labels to define several clustering metrics.  See Appendix \ref{sec:eval_metrics_formulas} for explicit formulas.  
\begin{enumerate}
    \item {\bf Adjusted Rand Index (ARI):}  The adjusted Rand index \cite{steinley2004properties} counts the proportion of data points that are clustered consistently with the true labels after adjusting for random chance.  It takes values in the range of $[-1,1]$ with value $1$ being a perfect clustering, $0$ being entirely random, and negative values being worse than random.  
    \item {\bf Normalized Mutual Information (NMI):}  The normalized mutual information \cite{mcdaid2011normalized} is an information-theoretic metric which quantifies how much the knowledge of the predicted labels reduces the information content of the true labels.  It is normalized to take values in $[0, 1]$ with $1$ being perfect correspondence and $0$ indicating that the two sets of labels are independent.  
    \item {\bf Purity:}  Purity \cite{amigo2009comparison} measures the homogeneity of the predicted clusters in terms of the true cluster assignments.  
    In the setting of two predicted clusters and two true clusters, the purity is just the accuracy of the clustering.
\end{enumerate}
\subsection{Comparators}
\label{sec:comparators}
We tested the method against several benchmark methods. We implemented each method in a manner similar to that of CLOCK. Thus, each implementation consisted of the following general steps: 
\begin{enumerate}[label = (\alph*)]
\item Map the time series to representative feature vectors in latent space (using the identity mapping by default).  
\item Compute pairwise distances between the feature vectors using a distance metric (using Euclidean distance by default).
\item Convert distances to similarities using a Gaussian kernel, choosing the median $\sigma_d := \mbox{\rm median} \{d(y_i, y_j)\}_{i \neq j}$ as the bandwidth.
\item Obtain cluster assignments via spectral clustering on the matrix of similarities.
\item By default, use linear interpolation to handle irregular spacing. 
\end{enumerate}
 We tested against the following collection of methods:
\begin{enumerate}
    \item {\bf Euclidean distance:}  This baseline method uses ordinary Euclidean distance to compute distances between time series.  If the time series $y_i$ are given as matrices with shape $T \times d$, with $T$ the number of timestamps and $d$ the dimension, we compute $d^2(y_i, y_j) = \|y_i - y_j\|^2_{\mathbb{R}^{T \times d}}$.  Thus, the similarity metric $S(\cdot, \cdot)$ is given by a standard Gaussian kernel.
    \item {\bf Dynamic Time Warping (DTW):} We compute the distance between time series $y_i$ and $y_j$ with the DTW algorithm \cite{1163055}. This algorithm aligns two time series by stretching and compressing segments with a dynamic one-to-many mapping before computing the distance between the aligned points.  Time series with similar features that are out of phase along the time axis will be nearby in terms of DTW distance.  
    \item {\bf Shape-Based Distance (SBD):}  SBD is a distance metric for time series used in the KShape algorithm \cite{kshape}.  Like DTW, it aligns the time series before computing distance.  However, unlike DTW, it aligns time series by shifting; not stretching and compressing.  The cross correlation statistic $c$, which is a similarity measure in the range $(-1, 1)$, of the aligned, $z$-normalized time series is computed,  and the distance is given by $1 - c$.  This distance is invariant to scaling and shifting.  
    \item {\bf Catch22:}  The Catch22 method \cite{lubba2019catch22} extracts 22 specially-chosen features from each time series to use as a low-dimensional representation.  The 22 selected features were chosen from a pool of over 7000 candidate features and found to be optimal across a broad range of tasks (such as clustering and classification) using rigorous evaluation criteria.  
    \item {\bf Latent ODE:}  This is a deep learning method \cite{rubanova2019latent} which models the time series as trajectories generated by a learned vector field and initial conditions in a latent, high-dimensional space.  Since dynamical information from the trajectories is encoded in the latent initial conditions, we used them as feature vectors for the time series.  This method takes irregular spacing into account, and thus we don't have to interpolate.  
    
     \item {\bf TS2Vec:}  TS2Vec \cite{yue2022ts2vec} is another deep learning method using a contrastive framework to learn representations of time series.  During training, positive pairs of context views are optimized via a contrastive loss.  The context views are aligned across the time dimension, as well as across spatial dimensions.  The contrastive loss is applied hierarchically at various scales to account for low- and high-resolution features.  
     To handle irregular spacing, we formed a uniform grid of time-points $[t_{min}, t_{max}]$ and inserted samples at times corresponding to observations.  We masked out the other points.  
\end{enumerate}
Latent ODE and TS2Vec have several hyperparameters to tune.  We used an internal clustering metric, the density-based clustering validation (DBCV) score, for validating them.  See Appendix \ref{sec:hyperparameter_tuning} for details.  See Table \ref{tab:results_table} for the results of the numerical experiments.  

\subsection{Ablation studies}
\label{sec:ablation_studies}
We carried out two ablation studies to determine the the rate of convergence of time-averaged occupation kernels to the associated KMEs of SRB measures, and the quality of the clustering of the CLOCK method as a function of trajectory length.  For both experiments, we used the Gaussian uniscale kernel (see section \ref{sec:kernels_used} for details).

\begin{figure*}
    \begin{subfigure}{0.45\linewidth}
        \includegraphics[width=1.0\linewidth]{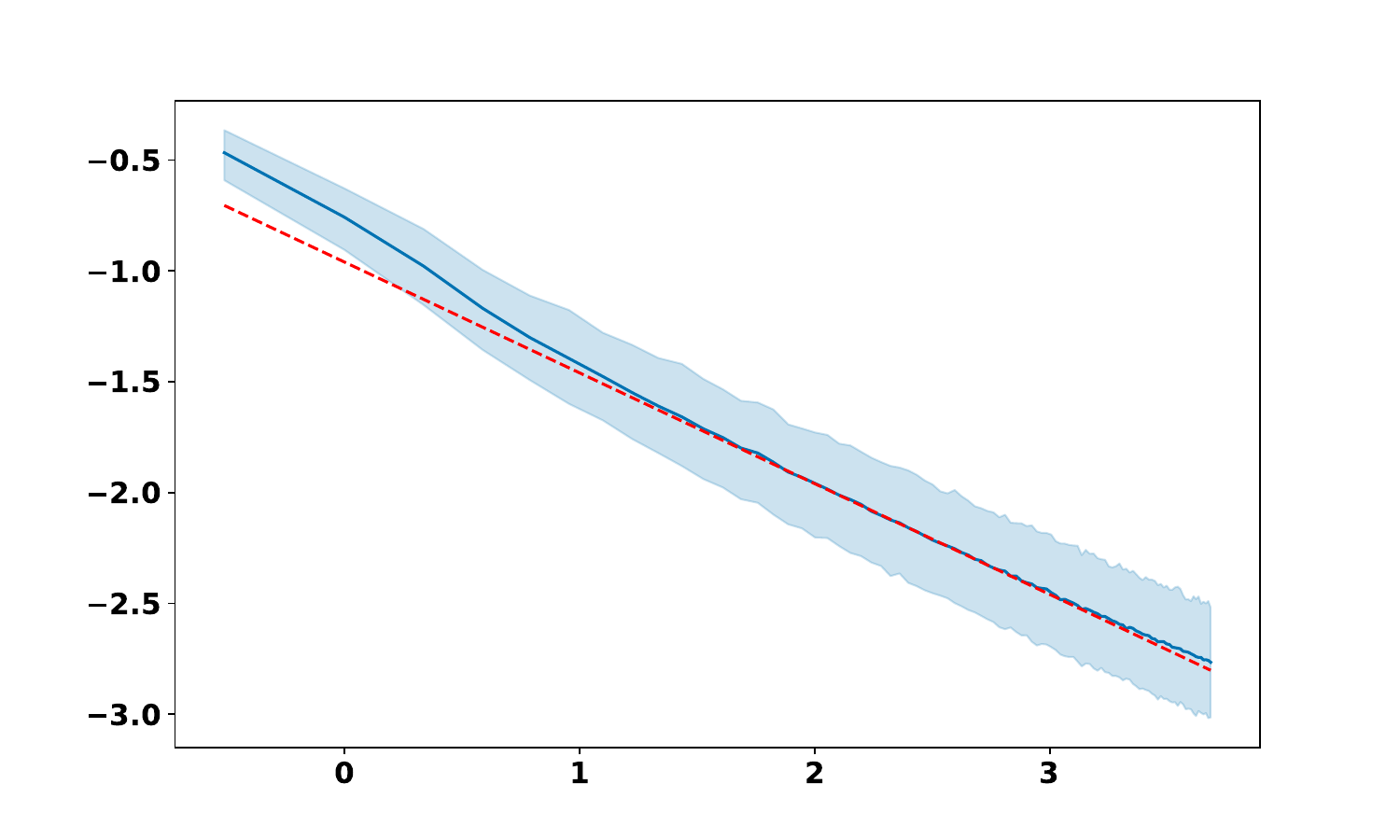}
        \caption{}
    \end{subfigure}
    \begin{subfigure}{0.45\linewidth}
        \includegraphics[width=1.0\linewidth]{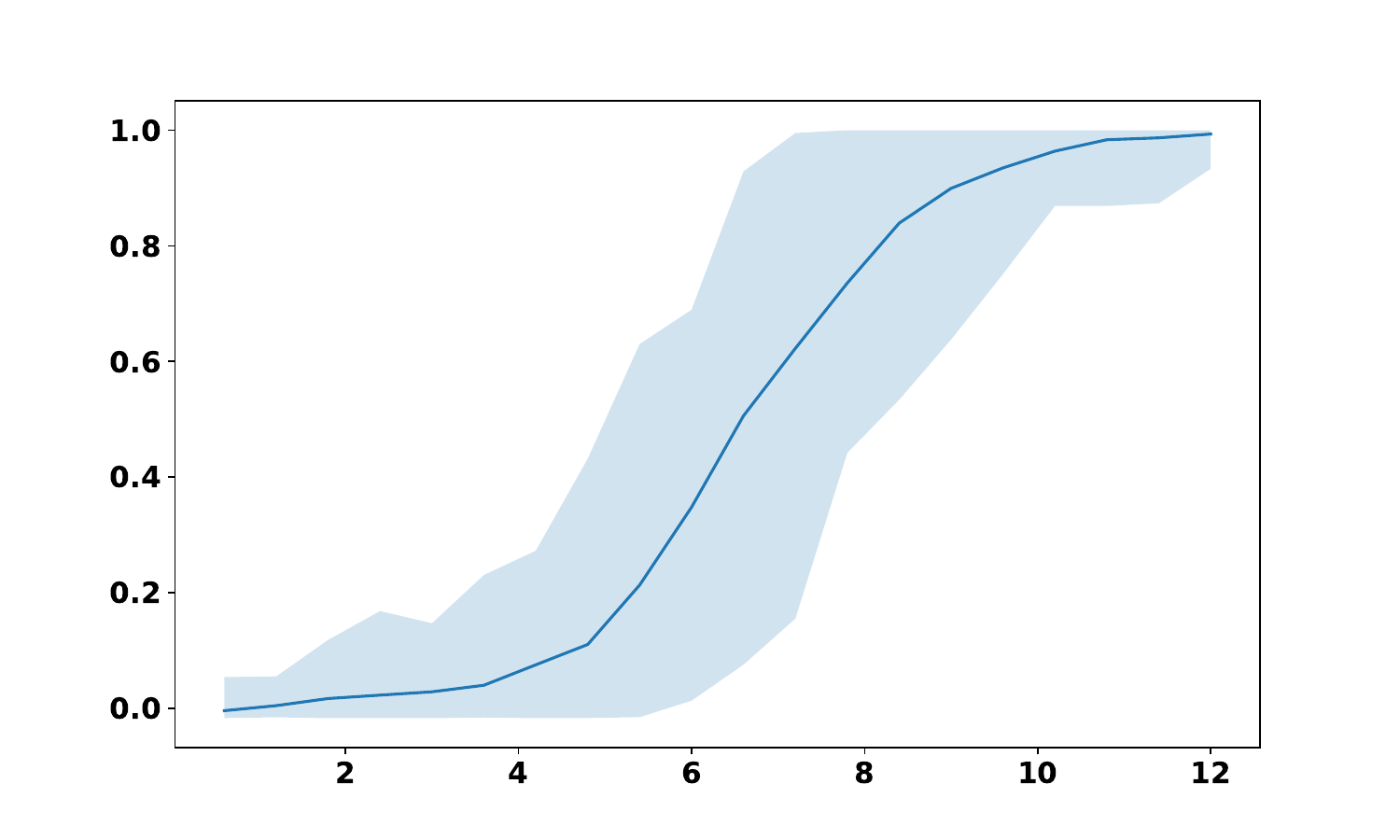}
        \caption{}
    \end{subfigure}
    \caption{In (a), the log-mean-squared error between the time-averaged occupation kernels and the empirical KME of the SRB measure of the Lorenz-96 attractor is plotted as a function of the log of the trajectory length $T$, along with a 95\% confidence interval (CI).  
    The line with slope $-0.5$ and intercept $-0.96$ is plotted for reference.
    In (b), the ARI of clusterings with the CLOCK method on the Lorenz-96 system is plotted with a 95\% CI as a function of $T$.}
    \label{fig:ablation_figures}
\end{figure*}

The convergence rate of time-averaged integrals of observables over trajectories to the spatial integral of the observable with respect to the associated SRB measure is guaranteed to be $\mathcal{O}(T^{-1/2})$ for systems with sufficient regularity (see \cite{young1999recurrence} for details).  We computed the empirical convergence rate for the 40D Lorenz-96 system with $F=8$.  To do so, we integrated a single trajectory over the interval $[0, 300]$ with random initial condition in the box $[-10,10]^{40}$.  We computed a time-averaged occupation kernel for this trajectory, and used it as a proxy for the KME of the SRB measure of the Lorenz-96 attractor, denoted $\hat{\mu}$.  For each of a collection of equally-spaced increasing time-points $\{T_j\}_{j=1}^{20}, T_j < 300$, we integrated 1000 trajectories over the interval $[0, T_j]$ with random initial conditions drawn from the box $[-10,10]^{40}$ and computed $\|L_{i,j}^* - \hat{\mu}\|_H$ for $i=1,\ldots, 1000, j=1,\ldots, 20$, where $L_{i,j}^*$ is the time-averaged occupation kernel of trajectory $i$ of length $T_j$.  A log-log plot of the resulting mean error over all trajectories as a function of trajectory length is plotted in Figure \ref{fig:ablation_figures} (a).  Note that after an initial "burn-in" period, we observe a straight line with slope $\approx -0.48$, as expected.

For the next experiment, we generated two sets of trajectories for the 40D Lorenz-96 system, each with 30 trajectories.  The first set corresponded to the parameter choice of $F=8$, and the second to $F=10$.  Each trajectory was integrated over the interval $[0,15]$ with initial conditions drawn randomly from the box $[-10, 10]^{40}$.  For each of an equally-spaced increasing sequence of time-points $\{T_j\}_{j=1}^{100}, 0 \leq T_j < 6$, we clustered the trajectories truncated to the interval $[0, T_j]$ with the CLOCK method.  The mean ARI over 40 runs as a function of trajectory length is plotted in Figure \ref{fig:ablation_figures} (b). 

\begin{table*}[htbp]
\small
  \centering
  \caption{Results of Numerical Experiments.  The CLOCK method is listed along with its accompanying kernel.   Information on the kernels used can be found in Appendix \ref{sec:kernels_used}.  The Gaussian Multiscale kernel (GM) with one bandwidth per dimension was the default choice.  The Gaussian Uniscale kernel (GU) with a single bandwidth for all dimensions was chosen for high-dimensional data.  The quadratic kernel (quad) and Laplace kernel were also used.  Best results for each metric are in bold.}  
\label{tab:results_table}
  \begin{subtable}{.4 \linewidth}
\begin{tabular}{l l l l l}
        \toprule
        \multicolumn{4}{c}{FHN}\\
        Method & ARI & NMI & Purity\\
        CLOCK GM & {\bf 1} & {\bf 1} & {\bf 1}\\
        CLOCK quad &  0.6893 &0.6615 & 0.9167\\
        Euclidean & 0.0638 & 0.1892 & 0.6333\\
        DTW & 0.0821 & 0.2102 & 0.65\\
        SBD & 0.1508 & 0.2746 & 0.7\\
        Catch22 & 0.634 & 0.619 & 0.9\\
        Latent ODE & 0.1209 & 0.1144 & 0.6833\\
        TS2Vec & {\bf 1} & {\bf 1} & {\bf 1}\\
        \bottomrule
    \end{tabular}%
     \end{subtable}
     \hspace{0.75cm}
       \begin{subtable}{.4 \linewidth}
\begin{tabular}{l l l l l}
        \toprule
        \multicolumn{4}{c}{Van der Pol}\\
        Method & ARI & NMI & Purity\\
        CLOCK GM & 0.6893 & 0.6615 &0.9167 \\
        CLOCK quad & -0.0172 &0 & 0.5\\
        Euclidean &-0.0172  &0  &0.5 \\
        DTW & 0 &0.0301  &0.5167 \\
        SBD & {\bf 1} & {\bf 1} & {\bf 1}\\
        Catch22 & 0.7469 & 0.7084 &0.9333 \\
        Latent ODE &-0.0124  & 0.0033&0.5333 \\
        TS2Vec & {\bf 1} & {\bf 1} & {\bf 1}\\
        \bottomrule
    \end{tabular}%
     \end{subtable}
            \begin{subtable}{.4 \linewidth}
\begin{tabular}{l l l l l}
        \toprule
        \multicolumn{4}{c}{Lorenz63}\\
        Method & ARI & NMI & Purity\\
        CLOCK GM & {\bf 1} & {\bf 1} & {\bf 1}\\
        CLOCK quad & 0.8689 &0.8214  & 0.9667 \\
        Euclidean &0.0969 & 0.0911 & 0.6667\\
        DTW & 0 & 0.0301 & 0.5167\\
        SBD &-0.0144  &0  &0.5 \\
        Catch22 & -0.0124 &0.0033  & 0.5333\\
        Latent ODE & {\bf 1} &{\bf 1}  &{\bf 1} \\
        TS2Vec & {\bf 1} &{\bf 1}  &{\bf 1} \\
        \bottomrule
    \end{tabular}%
     \end{subtable}
     \hspace{0.75cm}
                 \begin{subtable}{.4 \linewidth}
\begin{tabular}{l l l l l}
        \toprule
        \multicolumn{4}{c}{Kuramoto}\\
        Method & ARI & NMI & Purity\\
        CLOCK GM & {\bf 0.6893} & {\bf 0.6615} &{\bf 0.9167} \\
        CLOCK quad & 0.1197 &0.1008  &0.6833  \\
        Euclidean &0.0958 &0.0817  &0.6667 \\
        DTW & 0.1508 & 0.2746 & 0.7\\
        SBD & -0.0172 & 0 & 0.5\\
        Catch22 &-0.0126  &  0.0032&0.5333 \\
        Latent ODE &-0.007  & 0.0072&0.55\\
        TS2Vec & 0.0011 &0.0132& 0.5667  \\
        \bottomrule
    \end{tabular}%
     \end{subtable}
                 \begin{subtable}{.4 \linewidth}
\begin{tabular}{l l l l l}
        \toprule
        \multicolumn{4}{c}{Lorenz96}\\
        Method & ARI & NMI & Purity\\
        CLOCK GU & {\bf 1} & {\bf 1} & {\bf 1}\\
        CLOCK quad &  0.3922& 0.449 &0.8167  \\
        Euclidean &0 &0.0301  & 0.5167\\
        DTW &0  & 0.0301 & 0.5167\\
        SBD &-0.0105  & 0.0013 &0.5167 \\
        Catch22 & -0.0022 & 0 & 0.5\\
        Latent ODE & -0.0041 & 0.0037 & 0.5167\\
        TS2Vec & 0.8689 &0.8214  &0.9667 \\
        \bottomrule
    \end{tabular}%
     \end{subtable}
     \hspace{0.75cm}
       \begin{subtable}{.4 \linewidth}
\begin{tabular}{l l l l l}
        \toprule
        \multicolumn{4}{c}{EEG}\\
        Method & ARI & NMI & Purity\\
        CLOCK GM &  0.6873&0.5821  &0.915 \\
        CLOCK Laplace &{\bf 0.7383}   &{\bf 0.6341 }  & {\bf 0.93} \\
        Euclidean & 0.0006 &0.0273  &0.515  \\
        DTW  &0.1199  & 0.242 &0.675  \\
        SBD  &0.0604  & 0.1787 &  0.625\\
        Catch22 & 0.0183 &0.1081  &0.57 \\
        Latent ODE & 0.0421& 0.1246& 0.605\\
        TS2Vec &0.6874 &0.6159 & 0.915\\
        \bottomrule
    \end{tabular}%
     \end{subtable}
\end{table*}
\section{Discussion}
As can be seen from Table \ref{tab:results_table}, the CLOCK method achieved superior performance on all but one dataset.  The quadratic kernel failed on the Van der Pol dataset most likely due to the degree mismatch between the underlying system and the kernel.  However, no other method did as well as CLOCK overall, with the closest competitor being TS2Vec.  TS2Vec experienced a significant degradation of performance on the Kuramoto dataset.  This may be due to the trajectories in this dataset being locally similar but globally divergent (see Figure \ref{fig:kuramoto}), so that the hierarchical contrastive framework encoded dissimilar trajectories with similar feature vectors.  
We speculate that the overall good performance of the CLOCK method can be attributed to its capacity to accommodate the dynamical information contained in the trajectories.
Although Latent ODE also incorporates dynamical information, it yields a single dynamical model for trajectories with multiple dynamics, which may be problematic.
These results highlight the need for a method specifically-designed to cluster trajectories generated by differential equations.
In Appendix \ref{sec:noise_test}, we demonstrate empirically how the CLOCK algorithm maintains good performance on extremely noisy datasets.


\subsection{Limitations}
\label{sec:limitations}
    While the proposed embedding and clustering framework can be applied to general time series without explicit assumptions, its theoretical motivation relies on the setting where trajectories are generated by dynamical systems. In practice, there are datasets for which this assumption is not satisfied. In such cases, it may be necessary to first transform the data into a representation that better reflects an underlying dynamical structure, for instance via delay-coordinate embeddings such as Takens' embedding.
    
    The method assumes that the time series are of sufficient length to capture the nature of the invariant measures of the attractors. Moreover, there exist trajectories generated by different dynamical systems that exhibit distinct short-term (transient) behaviors but share the same asymptotic behavior, i.e., are associated with the same SRB measure. In such cases, the proposed method may fail to distinguish between them, as it clusters trajectories based on their long-term statistical properties rather than their transient dynamics.  


\subsection{Conclusion}
We have presented a clustering method using a novel dissimilarity metric for time series.  The metric is given by the squared RKHS norm between time-averaged occupation kernels.  The Palis conjecture and Theorem \ref{cor:strong_convergence} assert that this dissimilarity metric measures how distinct the associated SRB measures are, 
and thus is able to quantify the difference in the dynamics between two time series.

The method has displayed robust performance on a wide variety of synthetic datasets: low and high dimensions, regular and irregular spacing, and all with a significant amount of noise.  This suggests that the RKHS distance between time-averaged occupation kernels is an effective metric whenever it is suspected that the data was generated by differential equations.  In addition, this method achieved the best performance on a real dataset consisting of EEG measurements.  
\subsection{Future work}
There are a several directions for future research:  Spectral clustering may be replaced with K-Means or community clustering.  Additionally, different kernels may be selected to compute embeddings, and there are many ways to convert dissimilarities into similarities. A particularly promising direction is the extension of our framework to stochastic dynamical systems, including stochastic differential equations and continuous-time Markov chains where clustering could be performed based on invariant measures.

\subsection*{AI Use Statement}
\label{sec:LLM_disclosure}
We were assisted by the Gemini 3.1 Pro LLM model in this research.  Specifically, the LLM assisted in locating the foundational result that this work is based on: the Palis conjecture justifying the use of time-averaged occupation kernels as feature vectors for time series.  Additionally, the LLM assisted with the design of the numerical experiments by recommending datasets and evaluation metrics to use.  However, all theoretical proofs and numerical experiments were carried out by the authors themselves, and all theoretical results referenced by the LLM were verified in the original source material.  We take responsibility for the final content of this work.

    
\newpage
\printbibliography
\newpage

\onecolumn
\appendix
\aistatstitle{Supplementary Materials}
\section{Theoretical results}
\label{sec:theoretical_background}
An SRB measure $\mu$ for an attractor of an ODE defined on a compact manifold $M$ has the property that
\begin{equation}
    \label{eq:phys_measure}
    \lim_{T \to \infty}\frac{1}{T}\int_0^T \phi(y(t)) dt = \int_M \phi(p) d\mu(p)
\end{equation}
for every continuous function $\phi \colon M \to \RR$ and trajectory $y \colon [0, \infty) \to M$ such that $y(0) \in B(\mu)$, the basin of attraction for $\mu$.  The Palis conjecture states that, for a dense set in the set of all $C^r$ maps ($r \geq 1$) on $M$ with the $C^r$ topology, there exists a finite collection of SRB measures $\mu_1,\ldots, \mu_n$ such that $\cup_i B(\mu_i)$ has total Lebesgue measure.  That is, $\lambda(\cup_i B(\mu_i)) = \lambda(M)$ where $\lambda(\cdot)$ denotes the Lebesgue measure.

We wish to show that \eqref{eq:phys_measure} extends to feature vectors in an RKHS.  More specifically, we want to show that
\begin{equation}
    \lim_{T \to \infty}\frac{1}{T}\int_0^T K(\cdot, y(t)) dt = \int_M K(\cdot, p)d\mu(p).
\end{equation}
This is the content of Theorem \ref{cor:strong_convergence} below.  In Corollary ???, we show that the rate of convergence is $\mathcal{O}(T^{-1/2})$.
\begin{proposition}
    \label{prop:weak_convergence}
    Let $K \colon M \times M \to \RR$ be a continuous kernel defined on a compact manifold $M$.  Let $\{T_i\}_{i=1}^{\infty}$ be an increasing sequence of positive numbers such that $T_i \to \infty$.  Let $y \colon [0, \infty) \to M$ denote a trajectory generated by an ODE such that $y(0) \in B(\mu)$, the basin of attraction of an SRB measure $\mu$.  Define the linear functional $L_i \colon H \to \RR$ according to
    \begin{equation}
        L_i \phi = \frac{1}{T_i}\int_0^{T_i}\phi(y(t)) dt.
    \end{equation}
    Then each $L_i$ is bounded, and their Riesz representers, denoted $L_i^*$, converge weakly to the function $\hat{\mu}$ defined by
    \begin{equation}
        \hat{\mu} := \int_M K(\cdot, p)d\mu(p).
    \end{equation}
\end{proposition}
\begin{proof}
    The boundedness of the $L_i$ follows from the compactness of $M$ and the continuity of $K$ (see \cite{rielly2025mock} for details).  For the same reason, the operator
    \begin{equation}
        \phi \mapsto \int_M \phi(p) d\mu(p)
    \end{equation}
    is bounded, and its Riesz representer is $\hat{\mu}$.  The functions $L_i^*$ are given by
    \begin{equation}
        L_i^* = \frac{1}{T_i}\int_0^{T_i}K(\cdot, y(t)) dt.
    \end{equation}
    Because $K$ is continuous, every function $\phi$ in the corresponding RKHS is continuous.  Thus, since $\mu$ is an SRB measure, we get
    \begin{equation}
        \langle \phi, L_i^* \rangle_H = L_i \phi = \frac{1}{T_i}\int_0^{T_i}\phi(y(t)) dt \to \int_M \phi(p) d\mu(p) = \langle \phi, \hat{\mu} \rangle_H.
    \end{equation}
\end{proof}
\begin{lemma}
    \label{thm:norm_convergence}
    Let $L_i^*$ and $\hat{\mu}$ be as in Proposition \ref{prop:weak_convergence}.  Then the following holds:
    \begin{equation}
        \|L_i^*\|_H \to \|\hat{\mu}\|_H
    \end{equation}
as $i \to \infty$.
\end{lemma}
\begin{proof}
    We have
    \begin{equation}
        \|L_i^*\|_H^2 = \langle L_i^*, L_i^* \rangle_H = \frac{1}{T_i^2}\int_0^{T_i}\int_0^{T_i}K(y(s), y(t)) ds dt.
    \end{equation}
    Since $K$ is continuous and $M$ is compact, $K$ is a Mercer kernel.  Thus,
    \begin{equation}
        \frac{1}{T_i^2}\int_0^{T_i}\int_0^{T_i}K(y(s), y(t))dsdt = \frac{1}{T_i^2}\int_0^{T_i}\int_0^{T_i}\sum_{j=1}^{\infty}\lambda_j \phi_j(y(s))\phi_j(y(t)) ds dt
    \end{equation}
    for some collection of non-negative eigenvalues $\{\lambda_j\}$ and eigenfunctions $\{\phi_j\}$.  Let
\begin{equation}
    h_{i, N} := \frac{1}{T_i^2}\int_0^{T_i}\int_0^{T_i}\sum_{j=1}^N \lambda_j \phi_j(y(s))\phi_j(y(t)) ds dt = \sum_{j=1}^N \lambda_j \left(\frac{1}{T_i}\int_0^{T_i} \phi_j(y(t)) dt \right)^2.
\end{equation}
We have that for any fixed $N$, the sequence $h_{i,N}$ converges as $i \to \infty$ since $\mu$ is an SRB measure.  Likewise, for any fixed $i$, the sequence has a limit as $N \to \infty$ since the sequence of partial sums converges to the time-averaged double integral of the kernel over two copies of the trajectory (using the Dominated Convergence Theorem to pass the limit inside the integrals).  We have that
\begin{equation}
    \lim_{N \to \infty}\lim_{i \to \infty} h_{i, N} = \lim_{N \to \infty} \int \int \sum_{j=1}^N  \lambda_j\phi_j(p)\phi_j(q) d\mu(p) d\mu(q) = \int \int K(p, q)d\mu(p)d\mu(q) = \|\hat{\mu}\|^2_H,
\end{equation}
where in the second equality we used the Dominated Convergence Theorem with the bound
\begin{equation}
    \left \lvert \sum_{j=1}^N \lambda_j \phi_j(p)\phi_j(q)\right\rvert \leq \sqrt{K(p, p)K(q, q)}.
\end{equation}
We also have
\begin{equation}
    \lim_{i \to \infty}\lim_{N \to \infty} h_{i, N} = \lim_{i \to \infty}\frac{1}{T_i^2}\int_0^{T_i}\int_0^{T_i}K(y(s), y(t)) ds dt = \lim_{i \to \infty}\|L_i^*\|^2_H,
\end{equation}
using Dominated Convergence in the first equality.  Thus, we need to show that the order of the limits is exchangeable.  Let $h_i$ denote the limit $\lim_{N \to \infty} h_{i,N}$. We have
\begin{align}
    \lvert h_{i,N} - h_i\rvert &= \left\lvert \frac{1}{T_i^2}\int_0^{T_i}\int_0^{T_i} \sum_{j=N+1}^\infty \lambda_j \phi_j(y(s))\phi_j(y(t))ds dt\right\rvert\\
    &\leq \frac{1}{T_i^2}\int_0^{T_i}\int_0^{T_i}\max_{(y_1, y_2) \in M \times M}\left\lvert \sum_{j=N+1}^{\infty}\lambda_j \phi_j(y_1)\phi_j(y_2)\right\rvert ds dt \nonumber\\
    &= \max_{(y_1, y_2) \in M \times M}\left\lvert \sum_{j=N+1}^\infty \lambda_j \phi_j(y_1)\phi_j(y_2)\right\rvert \to 0, \nonumber
\end{align}
where on the last line we used the uniform convergence of the Mercer kernel on $M \times M$. The bound on $\lvert h_{i, N} - h_i\rvert$ does not depend on $i$ and converges to zero.  Thus, convergence of $h_{i, N}$ in $N$ is uniform, and we may apply the Moore-Osgood Theorem to show that the order of the limits is exchangable.  The result follows.
\end{proof}

\begin{theorem}
    \label{cor:strong_convergence}
    Let the setting of Proposition \ref{prop:weak_convergence} hold.  Then we have the following:
    \begin{equation}
        \label{eq:main_thm_eq}
        \frac{1}{T_i}\int_0^{T_i}K(\cdot, y(t))dt \to \int_M K(\cdot, p)d\mu(p)
    \end{equation}
    as $i \to \infty$.
\end{theorem}
\begin{proof}
    By Proposition \ref{prop:weak_convergence} and Lemma \ref{thm:norm_convergence}, we have
    \begin{equation}
        \|L_i^* - \hat{\mu}\|^2_H = \|L_i^*\|^2_H - 2 \langle L_i^*, \hat{\mu} \rangle_H + \|\hat{\mu}\|^2_H
        \to \|\hat{\mu}\|_H^2 - 2\|\hat{\mu}\|^2_H + \|\hat{\mu}\|^2_H
        = 0.
    \end{equation}
\end{proof}
For systems with sufficient regularity (for example, hyperbolic systems; see \cite{young1999recurrence} for details), the time-average of an observable over a trajectory converges to the spatial average with respect to the associated SRB measure at a rate of $\mathcal{O}_{\mathbb{P}}(T^{-1/2})$.  We use this fact to show that the time-averaged occupation kernels with random initial conditions drawn from Lebesgue-absolutely-continuous distributions converge (in probability) to the KMEs of SRB measures at the same rate.

\begin{definition}
\label{def:auto_corr}
Let $f \colon M \to M$ be a $C^r$ ($r \geq 1$) map on $M$.  Let $\Phi_t \colon M \to M$ be the flow of $f$ at time $t$.  Define the {\em centered-kernel function} $\overline{K} \colon M \times M \to \mathbb{R}$ according to:
\begin{equation}
    \overline{K}(x, y) := K(x, y) - \int_M K(x, p)d\mu(p) - \int_M K(q, y)d\mu(q) + \int_M\int_M K(p, q) d\mu(p)d\mu(q).
\end{equation}
Define the {\em auto-correlation function for $\overline{K}$} as:
\begin{equation}
   C_{\overline{K}}(\tau) := \int_M \overline{K}(p, \Phi_{\tau}(p)) d\mu(p)
\end{equation}
\end{definition}
First, we prove a convergence result for the case that the initial condition is distributed according to $\mu$, an SRB measure for $f$.  We then extend the result to more general measures in the next result.
\begin{lemma}
\label{lem:big_lemma}
Assume we are in the setting of Definition \ref{def:auto_corr}.    Let $X \sim \mu$, where $\mu$ is an SRB measure for $f$.  Suppose that the following regularity condition holds:
\begin{equation}
    \sigma_{\overline{K}}^2 := 2 \int_0^\infty \lvert C_{\overline{K}}(\tau)\rvert d\tau < \infty
\end{equation}
Then we have the following convergence in probability:
\begin{equation}
    \left\|\frac{1}{T}\int_0^T K(\cdot, \Phi_t(X)) dt - \hat{\mu}\right\|_H \to_{\mathbb{P}} 0
\end{equation}
at the rate $\mathcal{O}_{\mathbb{P}}(T^{-1/2})$.
\end{lemma}
\begin{proof}
Denote by $L^*(\mu, T)$ the following:
\begin{equation}
    L^*(\mu, T) := \frac{1}{T}\int_0^T K(\cdot, \Phi_t(X)) dt 
\end{equation}
where $X \sim \mu$.  We have
\begin{equation}
    \label{eq:RKHS_norm_squared}
    \|L^*(\mu, T) - \hat{\mu}\|^2_H = \frac{1}{T^2}\int_0^T \int_0^T \overline{K}(\Phi_t(X), \Phi_s(X)) ds dt.
\end{equation}
where $\overline{K}$ is the centered kernel in Definition \ref{def:auto_corr}.
We have
\begin{equation}
    \mathbb{E}_{\mu}\left[\left\|L^*(\mu, T) - \hat{\mu}\right\|^2_H \right] = \frac{1}{T^2}\int_0^T \int_0^T C_{\overline{K}}(\lvert t - s\rvert)ds dt. 
\end{equation}
by integrating the right side of \eqref{eq:RKHS_norm_squared} with respect to $\mu$ over $M$ and using the $f$-invariance of $\mu$. Taking the change of variables $\tau = t - s$, we get
\begin{equation}
     \mathbb{E}_{\mu}\left[\left\|L^*(\mu, T) - \hat{\mu}\right\|^2_H \right] = \frac{2}{T}\int_0^T\left(1 - \frac{\tau}{T}\right) C_{\overline{K}}(\tau) d\tau \leq \frac{2}{T} \int_0^T \lvert C_{\overline{K}}(\tau)\rvert d\tau.
\end{equation}
Using Markov's inequality in conjunction with the regularity condition, we get that
\begin{equation}
    \mathbb{P}(\|L^*(\mu, T) - \hat{\mu}\|_H \geq \delta) \leq \frac{\sigma_{\overline{K}}^2}{\delta^2 T}.
\end{equation}
Set $\eta := \frac{\sigma_{\overline{K}}^2}{\delta^2 T}$.  Then, with probability at least $1 - \eta$, we have
\begin{equation}
    \|L^*(\mu, T) - \hat{\mu}\|_H \leq \frac{\sigma_{\overline{K}}}{\sqrt{\eta T}},
\end{equation}
giving us a convergence rate of $\mathcal{O}_{\mathbb{P}}(T^{-1/2})$.
\end{proof}
Suppose that the system $f$ is exponentially mixing, meaning that correlations decay exponentially \cite{young1998statistical}.  Then the auto-correlation function of the centered kernel will be integrable for kernels which are $\alpha$-H\"{o}lder continuous, for $\alpha \in (0, 1]$.  This holds since, in this case, $\lvert C_{\overline{K}}(t)\rvert \leq Ce^{-\gamma t}\|K\|_{C^\alpha}$ \cite{dolgopyat1998decay}, and so $2\int_0^\infty \lvert C_{\overline{K}}(\tau)\rvert d\tau < \infty$.  Standard kernels such as Gaussian, Laplace, Mat\'{e}rn, and polynomial satisfy this condition.

\paragraph{Setting.}
Let $f$ be a $C^1$ vector field on $\mathbb{R}^d$ with flow $\Phi_t$, and let
$M \subset \mathbb{R}^d$ be compact and forward invariant, i.e.\
$\Phi_t(M) \subset M$ for all $t \geq 0$ (e.g.\ a trapping region containing
the attractor). Let $\mu$ be a $\Phi$-invariant Borel probability measure on
$M$, let $K$ be a continuous positive-definite kernel on $M$ with RKHS $H$,
and write $K_x := K(\cdot, x)$ and $\hat{\mu} := \int_M K_p \, d\mu(p)$. For
$x \in M$ and $T > 0$ define
\begin{equation}
    L_T(x) := \frac{1}{T}\int_0^T K_{\Phi_t(x)}\, dt \in H .
\end{equation}
Since $\Phi$ and $x \mapsto K_x$ are continuous, $x \mapsto L_T(x)$ is
continuous from $M$ to $H$; in particular it is Borel measurable.
 
\paragraph{Hypotheses.}
\begin{enumerate}
    \item[(\textbf{K})] \emph{H\"older feature map.} There exist $L_K > 0$
    and $\kappa \in (0,1]$ such that
    $\|K_x - K_y\|_H \leq L_K |x - y|^{\kappa}$ for all $x, y \in M$.
    \item[(\textbf{L})] \emph{Rate under the invariant measure.} There exist
    $c \geq 0$ and $T_0 > 0$ such that, for $\tilde{X} \sim \mu$,
    \begin{equation}
        \mathbb{E}\left[\|L_T(\tilde{X}) - \hat{\mu}\|_H^2\right]
        \leq \frac{c}{T} \qquad \text{for all } T \geq T_0 .
    \end{equation}
\end{enumerate}
 
\begin{assumption}[Asymptotic phase with absolutely continuous phase distribution]
\label{ass:asymptotic_phase}
Let $X \sim \nu$, where $\nu$ is a Borel probability measure on $M$. There
exist a random variable $Y$ with values in $M$, defined on a common
probability space with $X$, a constant $\lambda > 0$, and a random variable
$C(X, Y)$ that is finite almost surely, such that
\begin{enumerate}[label=(A\arabic*)]
    \item $|\Phi_t(X) - \Phi_t(Y)| \leq C(X, Y)\, e^{-\lambda t}$ for all
    $t \geq 0$, almost surely;
    \item the law of $Y$ is absolutely continuous with respect to $\mu$.
\end{enumerate}
\end{assumption}
 
Assumption~\ref{ass:asymptotic_phase} says that the orbit of a typical initial
condition is exponentially asymptotic to the orbit of a point $Y$ on the
attractor (an \emph{asymptotic phase} of $X$), and that the law of $Y$ is not
concentrated on a $\mu$-null set. For hyperbolic attractors this is the
mechanism by which SRB measures are physical.
 
\begin{theorem}
\label{thm:major_thm_2}
Assume (\textbf{K}), (\textbf{L}), and Assumption~\ref{ass:asymptotic_phase}.
Then
\begin{equation}
    \left\| \frac{1}{T}\int_0^T K(\cdot, \Phi_t(X))\, dt - \hat{\mu} \right\|_H
    = \mathcal{O}_{\mathbb{P}}\!\left(T^{-1/2}\right).
\end{equation}
More precisely, for every $\varepsilon > 0$ there exists $D_\varepsilon < \infty$
such that
\begin{equation}
    \mathbb{P}\left(\|L_T(X) - \hat{\mu}\|_H > D_\varepsilon T^{-1/2}\right) \leq \varepsilon
    \qquad \text{for all } T \geq \max(1, T_0).
\end{equation}
\end{theorem}
 
\begin{proof}
By the triangle inequality,
\begin{equation}
    \|L_T(X) - \hat{\mu}\|_H
    \leq \underbrace{\|L_T(X) - L_T(Y)\|_H}_{\text{(I)}}
    + \underbrace{\|L_T(Y) - \hat{\mu}\|_H}_{\text{(II)}} .
\end{equation}
 
\emph{Term (I).} By (\textbf{K}) and (A1), almost surely,
\begin{align}
    \text{(I)} &\leq \frac{1}{T}\int_0^T \|K_{\Phi_t(X)} - K_{\Phi_t(Y)}\|_H\, dt
    \leq \frac{L_K\, C(X,Y)^{\kappa}}{T}\int_0^T e^{-\kappa\lambda t}\, dt \nonumber \\
    &\leq \frac{L_K\, C(X,Y)^{\kappa}}{\kappa \lambda T} .
\end{align}
 
\emph{Term (II).} Let $\rho := d\,\mathrm{Law}(Y)/d\mu$, which exists by (A2)
and is finite $\mu$-almost everywhere. For $\delta > 0$ let
$A_\delta := \{ y \in M : \|L_T(y) - \hat{\mu}\|_H > \delta \}$, a Borel set.
By Chebyshev's inequality and (\textbf{L}),
$\mu(A_\delta) \leq c/(\delta^2 T)$ for $T \geq T_0$. Hence, for any $B > 0$,
\begin{equation}
    \mathbb{P}(\text{(II)} > \delta)
    = \int_{A_\delta} \rho \, d\mu
    \leq \int_{\{\rho > B\}} \rho \, d\mu + B\, \mu(A_\delta)
    \leq \mathbb{P}\big(\rho(Y) > B\big) + \frac{B c}{\delta^2 T} .
\end{equation}
 
\emph{Conclusion.} Fix $\varepsilon > 0$. Since $\rho(Y) < \infty$ and
$C(X,Y) < \infty$ almost surely, choose $B$ with
$\mathbb{P}(\rho(Y) > B) \leq \varepsilon/3$ and $b$ with
$\mathbb{P}(C(X,Y) > b) \leq \varepsilon/3$. Set $a := \sqrt{3Bc/\varepsilon}$.
Taking $\delta = a T^{-1/2}$ above gives
$\mathbb{P}(\text{(II)} > a T^{-1/2}) \leq 2\varepsilon/3$. On the event
$\{C(X,Y) \leq b\}$ and for $T \geq 1$, term (I) is at most
$L_K b^{\kappa}/(\kappa\lambda T) \leq L_K b^{\kappa}/(\kappa\lambda\sqrt{T})$.
With $D_\varepsilon := a + L_K b^{\kappa}/(\kappa\lambda)$, a union bound
gives $\mathbb{P}(\|L_T(X) - \hat{\mu}\|_H > D_\varepsilon T^{-1/2}) \leq \varepsilon$
for all $T \geq \max(1, T_0)$.
\end{proof}
 
\begin{remark}[Sufficient conditions for (\textbf{L})]
\label{rem:condition_L}
The proof of Lemma~\ref{lem:big_lemma} establishes
$\mathbb{E}_{\mu}\|L_T(\tilde{X}) - \hat{\mu}\|_H^2 \leq \sigma_{\overline{K}}^2/T$,
so (\textbf{L}) holds with $c = \sigma_{\overline{K}}^2$ whenever
$\sigma_{\overline{K}}^2 < \infty$. A weaker sufficient condition is that the
primitive $S(\tau) := \int_0^\tau C_{\overline{K}}(s)\, ds$ is bounded,
$\sup_{\tau \geq 0}|S(\tau)| \leq \bar{S}$: integrating by parts in the proof
of Lemma~\ref{lem:big_lemma},
\begin{equation}
    \mathbb{E}_{\mu}\|L_T(\tilde{X}) - \hat{\mu}\|_H^2
    = \frac{2}{T}\int_0^T \Big(1 - \frac{\tau}{T}\Big) C_{\overline{K}}(\tau)\, d\tau
    = \frac{2}{T^2}\int_0^T S(\tau)\, d\tau
    \leq \frac{2\bar{S}}{T} .
\end{equation}
This weaker condition covers periodic attractors, for which
$C_{\overline{K}}$ is periodic and $\sigma_{\overline{K}}^2 = \infty$.
\end{remark}
 
\subsection{Verification for the kernels and systems of Section~\ref{sec:datasets}}
\label{sec:applicability}
 
\paragraph{Kernels.}
All kernels of Appendix~\ref{sec:kernels_used} satisfy (\textbf{K}). This
follows from $\|K_x - K_y\|_H^2 = K(x,x) + K(y,y) - 2K(x,y)$ together with
$1 - e^{-u} \leq u$ for $u \geq 0$, and, for the quadratic kernel, from its
explicit feature map $x \mapsto (1, \sqrt{2}\,x, x x^\top)$. With
$R := \max_{x \in M}|x|$:
\begin{itemize}
    \item GU with bandwidth $\eta$: $L_K = 1/\eta$, $\kappa = 1$;
    \item GM with bandwidths $\sigma_i$: $L_K = 1/\min_i \sigma_i$, $\kappa = 1$;
    \item Laplace with bandwidths $\sigma_i$: $L_K = (2\sqrt{d}/\min_i \sigma_i)^{1/2}$, $\kappa = 1/2$;
    \item quadratic: $L_K = \sqrt{2} + 2R$, $\kappa = 1$.
\end{itemize}
The bandwidths are computed once from the data and then held fixed; the
results apply conditionally on them.
 
\paragraph{FitzHugh--Nagumo.} For both parameter sets the unique equilibrium
$p$ is a hyperbolic sink (Jacobian eigenvalues $-0.104 \pm 0.28i$ and
$\{-3.46, -0.089\}$), and numerically all initial conditions in $[-2,2]^2$
converge to it. Hence $\mu = \delta_p$ and $L_T(p) = \hat{\mu}$, so
(\textbf{L}) holds with $c = 0$. Convergence to $p$ is exponential and
uniform on compact subsets of the basin, so
Assumption~\ref{ass:asymptotic_phase} holds with $Y := p$.
 
\paragraph{Van der Pol.} For $\mu > 0$ there is a unique hyperbolic attracting
limit cycle $\Gamma$, of period $P$, attracting every point except the origin.
For $y \in \Gamma$, each full period contributes exactly $P\hat{\mu}$, so
$\|L_T(y) - \hat{\mu}\|_H \leq 2P\bar{\kappa}/T$ with
$\bar{\kappa} := \sup_{x}\|K_x\|_H$, and (\textbf{L}) holds with
$c = 4P^2\bar{\kappa}^2$. (Here $C_{\overline{K}}$ is periodic, so
$\sigma_{\overline{K}}^2 = \infty$ and Lemma~\ref{lem:big_lemma} does not
apply as stated; cf.\ Remark~\ref{rem:condition_L}.) Take $Y$ to be the
asymptotic phase of $X$ on $\Gamma$: (A1) holds by hyperbolicity of $\Gamma$,
and since the asymptotic-phase map is a $C^1$ submersion on the basin \cite{guckenheimer1975isochrons}, (A2) holds when
$\nu$ has a density.

\section{Approximating the dissimilarity measure}
\label{sec:approx_dissimilarity}
We use integral quadratures to approximate the integrals in the dissimilarity measure in \ref{eq:distance_metric}. Assuming $m_i=m_j=m$, 
\begin{multline}
    \label{eq:distance_metric_discrete}
    \hat{d}^2(y_i, y_j) = 
    \frac{1}{T_i^2} \sum_{k=0}^{m} \sum_{l=0}^{m} w_{ik} w_{il} K(y_i(t_{ik}), y_i(t_{il})) + 
    \frac{1}{T_j^2} \sum_{k=0}^{m} \sum_{l=0}^{m} w_{jk} w_{jl} K(y_j(t_{jk}), y_j(t_{jl})) - \\ 
    \frac{2}{T_i T_j} \sum_{k=0}^{m} \sum_{l=0}^{m} w_{ik} w_{jl} K(y_i(t_{ik}), y_j(t_{jl})),
\end{multline}
The trapezoidal quadrature weights $w_{ik}$ for the $i$-th time series (and similarly for the $j$-th one) are defined as:
\begin{equation}
    w_{ik} = 
    \begin{cases} 
        \frac{t_{i,1} - t_{i,0}}{2} & \text{if } k = 0, \\[6pt]
        \frac{t_{i,k+1} - t_{i,k-1}}{2} & \text{if } 0 < k < m, \\[6pt]
        \frac{t_{i,m} - t_{i,m-1}}{2} & \text{if } k = m.
    \end{cases}
\end{equation}
If the time series is sampled uniformly such that $\Delta t_i = \frac{T_i}{m}$, the weights simplify to:
\begin{equation}
    w_{ik} = 
    \begin{cases} 
        \frac{\Delta t_i}{2} & \text{if } k = 0 \text{ or } k = m, \\[6pt]
        \Delta t_i & \text{if } 0 < k < m.
    \end{cases}
\end{equation}
\section{Kernels used}
\label{sec:kernels_used}
We used several different kernels in the numerical experiments in Section \ref{sec:experiments}.  We list their definitions here.
\begin{enumerate}
    \item {\bf Gaussian uniscale (GU):}
    \begin{equation}
        K(x, y) = \exp\left(-\frac{1}{2\eta^2}\|x-y\|^2\right),
    \end{equation}
    where we set $\eta := \mbox{\rm median}\{\|x_k - x_l\|\}_{k \neq l}$ where $x_k$ ranged over each individual observation in the dataset. This kernel was used for high-dimensional data.
    \item {\bf Gaussian multiscale (GM):}
    \begin{equation}
        K(x, y) = \exp\left(-\sum_{i=1}^d \frac{(x_i - y_i)^2}{2\sigma_i^2}\right),
    \end{equation}
    where $\sigma_i$ was set to be the standard deviation of the $i^{th}$ coordinate of the dataset.  This is the default kernel for CLOCK.
     \item {\bf Laplace:}
    \begin{equation}
        K(x,y) = \exp\left(-\sum_{i=1}^d \frac{\lvert x_i - y_i\rvert}{\sigma_i}\right),
    \end{equation}
    where $\sigma_i$ was set to be the standard deviation of the $i^{th}$ coordinate of the dataset.
    \item {\bf Quadratic (quad):}
    \begin{equation}
        K(x, y) = (x^Ty + 1)^2.
    \end{equation}
   
\end{enumerate}
\section{Formulas for evaluation metrics}
\label{sec:eval_metrics_formulas}
We provide explicit formulas for the evaluation metrics in Section \ref{sec:eval_metrics}.
\begin{enumerate}
    \item {\bf Adjusted Rand Index (ARI):}  The formula is given by
\begin{equation}
 \text{ARI} = \frac{ \sum_{i,j} \binom{n_{ij}}{2} - \frac{ \sum_{i} \binom{a_i}{2} \sum_{j} \binom{b_j}{2} }{ \binom{n}{2} } }{ \frac{1}{2} \left( \sum_{i} \binom{a_i}{2} + \sum_{j} \binom{b_j}{2} \right) - \frac{ \sum_{i} \binom{a_i}{2} \sum_{j} \binom{b_j}{2} }{ \binom{n}{2} } },
    \end{equation}
    where $n$ is the total number of time series in the dataset, $n_{ij}$ is the number of trajectories that are in both the true cluster $i$ and the predicted cluster $j$, $a_i$ is the total number of samples in true cluster $i$, and $b_j$ is the total number of samples in predicted cluster $j$.
    \item {\bf Normalized Mutual Information (NMI):}  The formula is given by
    \begin{equation}
        \text{NMI} = \frac{ 2 \sum_{i,j} n_{ij} \log \left( \frac{n \cdot n_{ij}}{a_i b_j} \right) }{ -\sum_{i} a_i \log \left( \frac{a_i}{n} \right) - \sum_{j} b_j \log \left( \frac{b_j}{n} \right) },
    \end{equation}
    where $n, n_{ij}, a_i, b_j$ are as above and by convention $0\log 0$ is set to $0$.
    \item {\bf Purity:}  If $\Omega$ is a predicted clustering and $C$ a true clustering, then the purity is defined as
    \begin{equation}
        \mbox{\rm purity}(\Omega, C) = \sum_{\omega \in \Omega} \max_{c \in C}\lvert \omega \cap c \rvert.
    \end{equation}
    In the setting of two predicted clusters and two true cluseters, the purity is just the accuracy of the clustering.
\end{enumerate}
\section{Hyperparameter tuning}
\label{sec:hyperparameter_tuning}
We validated the hyperparameters of the deep learning methods in Section \ref{sec:comparators} using an internal clustering metric, the DBCV.
\begin{enumerate}
    \item {\bf Density-based clustering validation (DBCV):}  The DBCV \cite{moulavi2014DBCV} is an internal clustering metric, meaning it can be used in the absence of true cluster labels.  
This metric assigns a high value to partitions with highly-dense clusters with sparse between-cluster regions.  It does so without assuming the clusters are globular, and so can measure the quality of clustering for highly-irregular collections of trajectories, which occurs frequently with chaotic systems, for example.  The DBCV score is normalized to be in the range $[-1, 1]$ with 1 being a perfect clustering and -1 being exceptionally poor.  We found it to be superior to the silhouette score \cite{shahapure2020cluster} for the task of validating hyperparameters.
\end{enumerate}
\section{Noise robustness experiment}
\label{sec:noise_test}
\begin{figure*}[ht!]
    \begin{subfigure}{0.47\linewidth}
\includegraphics[width=1.0\linewidth]{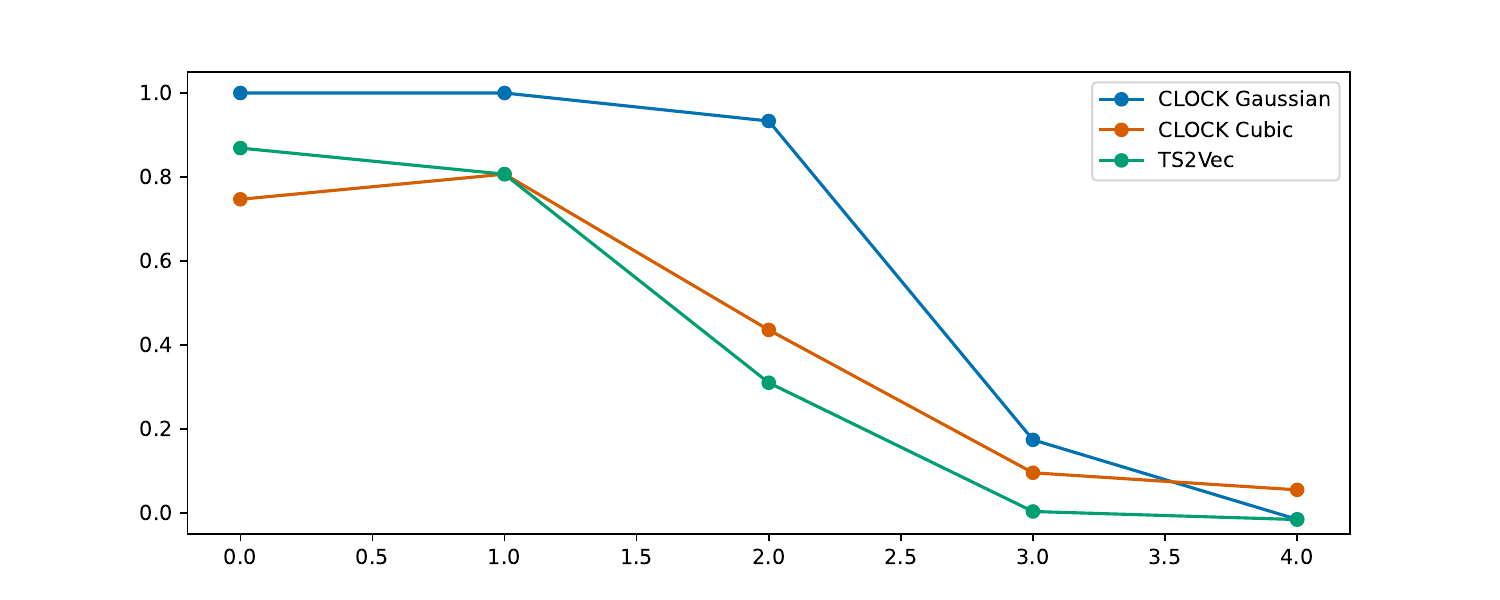}
    \caption{ARI}
    \label{fig:all_gauss}
    \end{subfigure}
    \begin{subfigure}{0.47\linewidth}
     \includegraphics[width=1.0\linewidth]{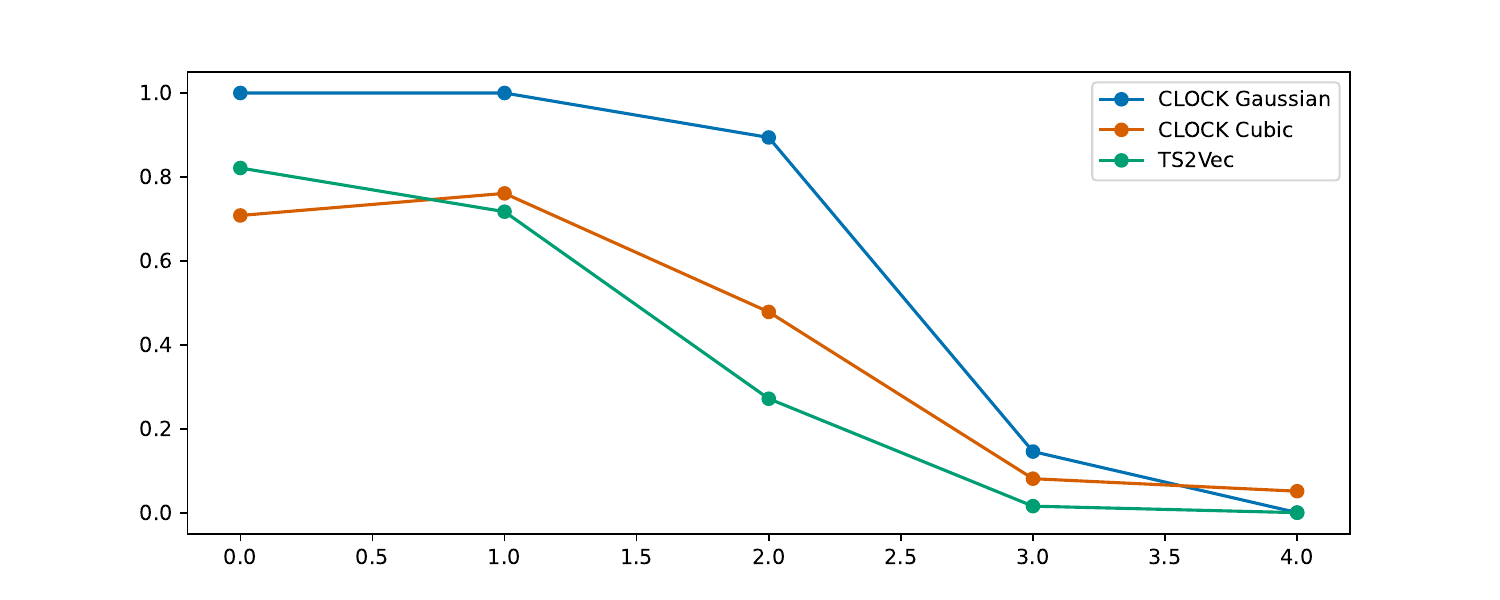}
    \caption{NMI}
    \label{fig:nmi_comparison}   
    \end{subfigure}
    \begin{subfigure}{0.47\linewidth}
         \includegraphics[width=1.0\linewidth]{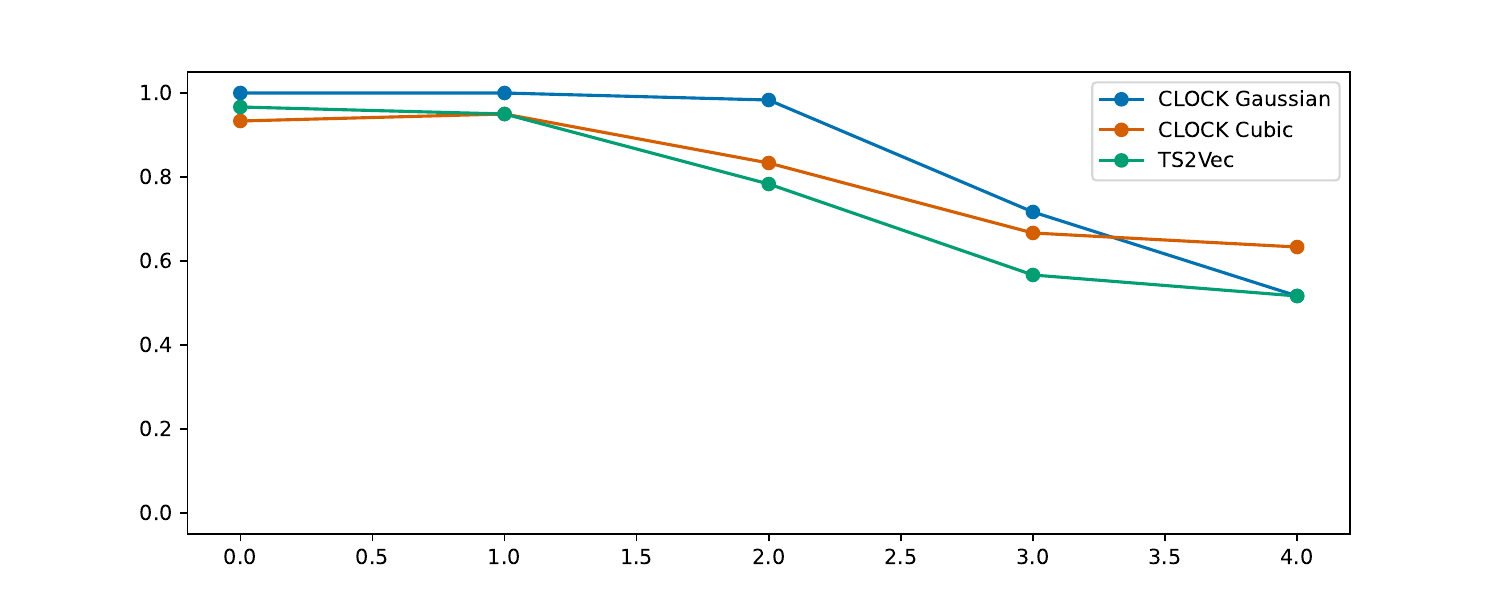}
    \caption{Purity}
    \label{fig:purity_comparison}   
    \end{subfigure}
    \caption{Plots of evaluation metrics in the noise robustness test of Appendix \ref{sec:noise_test}.  The horizontal axis represents the standard deviation of the Gaussian noise added to each observation.  The vertical axis represents one of the evaluation metrics: ARI, NMI or purity.  In Figure \ref{fig:all_gauss}, we have ARI results,  in Figure \ref{fig:nmi_comparison}, we have results for NMI, and in Figure \ref{fig:purity_comparison}, we have purity results.}
     \label{fig:noise_test_results}
\end{figure*}

We speculate that the CLOCK method is highly resistant to zero-mean Gaussian noise since the feature vector is obtained by performing an integration along a trajectory.  The integral effectively cancels out the perturbations caused by the noise.  In order to gauge the impact of noise on the performance of CLOCK, we performed an experiment in which we gradually increased the noise level in the Lorenz96 dataset while leaving the true trajectories fixed.  The CLOCK method with Gaussian uniscale and quadratic kernels (see Appendix \ref{sec:kernels_used}) was fit on each dataset, and we evaluated the quality of the clusterings.  We ran this experiment with the baseline method TS2Vec to use as a comparison.  The results are graphically represented in Figure \ref{fig:noise_test_results}.

The results of the noise robustness experiment show that CLOCK is resistant to noise.  It should be noted that the amount of noise added to the trajectories was extreme, even on the lower settings.  In Figure \ref{fig:noise_test_results}, we see that CLOCK with Gaussian kernel fails as the noise level approaches 3.0, while CLOCK with quadratic kernel is affected at level 2.0.  Thus, the choice of kernel is an important aspect of this clustering method.  In practice, the kernel may be treated as a hyperparameter and validated using an internal clustering metric.  TS2Vec generally performed worse than CLOCK with either kernel.

\section{Assets used}
\label{sec:licenses_for_assets}
The assets we used can be divided into methods, back-ends, and datasets.
\subsection{Methods}
\begin{enumerate}
    \item {\bf Dynamic Time Warping (DTW):}  We used the \texttt{dtw} function from the Python package \texttt{tslearn.metrics}.  This has the BSD-2-Clause license.  We used version 0.8.0.
    \item {\bf Shape-Based Distance (SBD):}  We used the package \texttt{scipy.signal} and the function \texttt{correlate}.  This has the BSD 3-Clause license.  We used version 1.17.1.
    \item {\bf Catch-22:} We used the package \texttt{pycatch22} and the function \texttt{catch22\_all}.  This has the GNU General Public License v3 (GPLv3) license.  We used version 0.4.5.
    \item {\bf Latent ODE:} We used the \texttt{pytorch} package and the \texttt{torchdiffeq} package.  The \texttt{pytorch} package has the BSD 3-Clause license.  We used version 2.10.0.  The \texttt{torchdiffeq} package has the MIT license.  We used version 0.2.5.
    \item {\bf TS2Vec:}  Code was downloaded from the GitHub repository at \url{https://github.com/zhihanyue/ts2vec}.  This has the MIT license.  We used the AAAI 2022 Official Release.
    \item {\bf Spectral clustering:}  We used \texttt{sklearn.cluster} package and the function \texttt{SpectralClustering}.  This has the BSD 3-Clause license.  We used version 1.8.0.
    \end{enumerate}
\subsection{Back-ends}
\begin{enumerate}
    \item {\bf JAX:} We used the \texttt{jax} and \texttt{jaxlib} Python packages.  These have the Apache 2.0 license.  We used version 4.30.0 of \texttt{jax} and version 4.30.0 of \texttt{jaxlib}.
\end{enumerate}
\subsection{Datasets}
\begin{enumerate}
    \item {\bf Bonn EEG dataset:}  Taken from the archive available at \url{https://www.upf.edu/web/ntsa/downloads/-/asset_publisher/xvT6E4pczrBw/content/2001-indications-of-nonlinear-deterministic-and-finite-dimensional-structures-in-time-series-of-brain-electrical-activity-dependence-on-recording-regi}.  This data is in the public domain.  
\end{enumerate}

\newpage

\end{document}